\documentclass[twoside,11pt]{article}
\usepackage{amsfonts}
\usepackage{mathrsfs}
\usepackage[utf8]{inputenc} % Use utf8
\usepackage{float}
\usepackage{enumitem}
\usepackage{amsmath,amssymb}
\usepackage{latexsym}
\usepackage{longtable}
\usepackage{makecell}
\usepackage{mathrsfs}

\usepackage{booktabs}
\usepackage{epstopdf}
\usepackage{tabularx}
\usepackage{tikz}
\usetikzlibrary{shapes.geometric, arrows}
\usetikzlibrary{positioning}
\tikzstyle{startstop} = [rectangle, rounded corners, minimum width=3cm, minimum height=1cm, text centered, draw=black, fill=red!30]
\tikzstyle{process} = [rectangle, minimum width=3cm, minimum height=1cm, text centered, draw=black, fill=blue!30]
\tikzstyle{io} = [trapezium, trapezium left angle=70, trapezium right angle=110, minimum width=3cm, minimum height=1cm, text centered, draw=black, fill=green!30]
\tikzstyle{arrow} = [thick,->,>=stealth]
\usepackage{geometry}  
\usepackage[active]{srcltx}
\usepackage{graphicx}
\usepackage{booktabs}
\usepackage{multirow}
\usepackage{siunitx}
\usepackage{xurl} % Allows URLs to break across lines
\usepackage{amsmath}
\usepackage{setspace}
\usepackage{graphicx} 
\usepackage{algorithm}
\usepackage{algpseudocode}

\usepackage{subcaption} % Required for subfigures
\usepackage[
    bookmarks=true,         % generates bookmarks for all entries in the "Table of Contents"
    bookmarksnumbered=true, % includes chapter-/header-/section-/subsection-/... -
    colorlinks=true, pdfstartview=FitV, linkcolor=blue, citecolor=blue,
    urlcolor=blue]{hyperref}

\usepackage{fancyhdr}
\usepackage{natbib}
\usepackage{lipsum} 
\usepackage{listings} % for displaying code
\usepackage{xcolor} % for setting colors
\definecolor{codegreen}{rgb}{0,0.6,0}
\definecolor{codegray}{rgb}{0.5,0.5,0.5}
\definecolor{codepurple}{rgb}{0.58,0,0.82}
\definecolor{backcolour}{rgb}{0.95,0.95,0.92}

\lstdefinestyle{mystyle}{
    backgroundcolor=\color{backcolour},
    commentstyle=\color{codegreen},
    keywordstyle=\color{magenta},
    numberstyle=\tiny\color{codegray},
    stringstyle=\color{codepurple},
    basicstyle=\footnotesize\ttfamily,
    breakatwhitespace=false,
    breaklines=true,
    captionpos=b,
    keepspaces=true,
    numbers=left,
    numbersep=5pt,
    showspaces=false,
    showstringspaces=false,
    showtabs=false,
    tabsize=2
}

\providecommand{\U}[1]{\protect\rule{.1in}{.1in}}

\usepackage{listings}
\usepackage{titlesec}
\titleformat{\section}
{\normalfont\Large\bfseries}{\thesection.}{1em}{}
\usepackage{fancyhdr}
\begin{document}
\date{}
\title{\Large \textbf{Copula Based Fusion of Clinical and Genomic Machine Learning Risk Scores for Breast Cancer Risk Stratification}}
\vspace{1ex}
\author{Agnideep Aich${ }^{1}$\thanks{Corresponding author: Agnideep Aich, \texttt{agnideep@stanford.edu}, ORCID: \href{https://orcid.org/0000-0003-4432-1140}{0000-0003-4432-1140}}
 \hspace{0pt}, Sameera Hewage${ }^{2}$ \hspace{0pt} and Md Monzur Murshed${ }^{3}$ 
\\ ${ }^{1}$ Department of Emergency Medicine, Stanford University, \\ Palo Alto, CA, USA. \\  ${ }^{2}$ Department of Mathematics, Southern Utah University, \\ Cedar City, Utah, USA\\  ${ }^{3}$ Department of Mathematics and Statistics, Minnesota State University, \\ Mankato, MN, USA
\\ }
\date{}
\maketitle

\vspace{-20pt}

\begin{abstract}
Clinical and gene-expression models are both used to predict breast cancer outcomes, but they are often combined using simple linear rules that do not account for how their risk scores relate. Using the METABRIC breast cancer cohort, we examined whether directly modeling the joint relationship between clinical and gene-expression machine-learning risk scores could improve risk stratification for 5-year cancer-specific mortality. We created a binary 5-year cancer-death outcome and defined two predictor views: a clinical view based on demographic, tumor, and treatment-related variables, and a gene-expression view based on mRNA expression features. We trained several supervised classifiers and used 5-fold cross-validated predicted probabilities as out-of-fold risk scores. These scores were converted to pseudo-observations on $(0,1)^2$ and used to fit Gaussian, Clayton, Gumbel, and Frank copulas. In the primary METABRIC analysis, the clinical model showed stronger discrimination (AUC 0.783) than the gene-expression model (AUC 0.721). Among the copula families considered, the Frank copula yielded the smallest goodness-of-fit statistic, while the Gaussian copula showed nearly identical performance. Although the copula-fused score did not improve ROC-AUC relative to the clinical model, joint risk stratification based on the clinical and gene-expression scores identified clear differences in long-term survival, with patients in the high-both group experiencing the least favorable outcomes. Competing-risks analysis showed the same overall pattern for the cumulative incidence of cancer death. In an independent TCGA cohort, we conducted an external evaluation under a reduced harmonized specification using shared
predictors and a common 5-year overall-mortality endpoint. The copula-fused score had discrimination comparable to the individual
and simple-fusion scores, with substantially overlapping confidence intervals. All three scores underwent the same METABRIC-based recalibration procedure before calibration was evaluated using the Brier score, calibration slope, calibration intercept, integrated
calibration index, Hosmer-Lemeshow diagnostic, and reliability curves. No gene met the prespecified stability criterion under repeated
cross-validated permutation importance; gene-level findings were therefore treated as exploratory biological context rather than
evidence of stable predictive drivers.
 We frame this work as a methodological and exploratory study of interpretable, dependence-aware score fusion: The results show that copula modeling provides an explicit description of dependence between clinical and gene-expression risk scores and supports descriptive joint score-group analyses. The study does not establish superior prediction, validated clinical risk
categories, or clinical utility.
\end{abstract}

\section*{Introduction}

Breast cancer is still a major cause of cancer deaths among women worldwide, with over two million new cases each year \cite{bray2018}. Data show that outcomes vary widely: five-year survival rates are high for localized cases but drop sharply for advanced or metastatic disease. Because of these differences, accurate risk assessment is needed to tailor treatments and focus intensive care on patients who are most likely to benefit.

Researchers now know that breast cancer is not a single disease but includes several biologically distinct subtypes. Gene-expression studies and subtype classifiers have identified groups, such as luminal A, luminal B, HER2-enriched, basal-like, and normal-like, each with unique molecular features and clinical outcomes \cite{parker2009}. Large projects, such as the Molecular Taxonomy of Breast Cancer International Consortium (METABRIC), have collected detailed clinical and gene-expression data from thousands of patients, providing a clearer picture of breast tumor biology \cite{curtis2012,pereira2016}. Still, even within the same subtype, there is considerable variation in patient outcomes, highlighting the need to combine different types of information.

Traditionally, breast cancer prognosis has been based on clinicopathologic factors like tumor size, grade, lymph-node status, hormone-receptor status, and HER2 expression. These markers are affordable, easy to collect, and straightforward to interpret, forming the basis of tools like the Nottingham Prognostic Index. However, clinical models alone may overlook important biological differences. Tumors that look alike under the microscope can behave very differently at the molecular level and respond differently to treatment.

Multigene tests like Oncotype DX, MammaPrint, and PAM50 have made gene-expression--based risk prediction a regular part of clinical care \cite{paik2004,van2002,parker2009}. These tests provide information on tumor growth, hormone signaling, and tumor subtypes that clinical markers alone cannot. However, studies show that gene-expression predictors do not consistently outperform clinical models, and their added prognostic value can vary across breast cancer subtypes and populations \cite{neapolitan2015integrated, liu2014breast}.
For example, in the METABRIC study, combining clinical and gene-expression data improved survival predictions; however, the relative contribution of clinical and gene-expression variables differed across subtypes, with each providing complementary information \cite{neapolitan2015integrated, liu2014breast}.

% However, studies show that gene-expression predictors do not always outperform clinical models, and their extra value can be limited at the population level \cite{neapolitan2015integrated,liu2014breast}. For example, in the METABRIC study, combining clinical and gene-expression data improved survival predictions; however, clinical factors still provided a wealth of useful information \cite{neapolitan2015integrated,liu2014breast}.

This suggests that clinical and gene-expression predictors capture both shared and unique aspects of risk.

Modern machine learning (ML) methods have introduced new approaches to modeling risk in complex cancer data. Flexible models, such as tree-based ensembles, can identify nonlinear effects, interactions, and complex relationships between genes and clinical factors. However, ML models can be influenced by specific features of the data and analysis choices, and reproducibility and generalisability remain significant concerns in oncology. 
% In many studies using datasets like METABRIC, clinical and gene-expression features are either combined into one large dataset or their risk scores are simply added together \cite{neapolitan2015integrated,liu2014breast}. 
In many studies using datasets like METABRIC, clinical and gene-expression features are either combined into one large dataset or their risk scores are simply added together, without explicitly accounting for subtype-specific differences in their prognostic contributions \cite{neapolitan2015integrated, liu2014breast}. These approaches do not necessarily provide a separate, explicit model of dependence between independently developed view 
specific scores. Combined-feature machine-learning models may capture nonlinearities and interactions implicitly, while simple linear
score combinations impose a more restricted functional structure. The copula approach differs by separating marginal prediction from
an explicit model of cross-view score dependence
% These approaches assume that the risk factors are either independent or simply add up in a straightforward manner.

% In reality, the impact of gene-expression information on prognosis can depend on a patient's clinical details, and vice versa. For instance, a gene-expression signature indicating high proliferation might have different implications for a young patient with node-positive disease compared to an older patient with small, node-negative tumours. To capture these context-specific effects, we need tools that can describe both the individual and joint behaviour of clinical and gene-expression risk scores. Standard methods, such as linear combination or simply merging features, cannot easily reveal complex dependencies or cases where risk is high only when both clinical and gene-expression factors are extreme.

Copula models provide a structured approach to separating individual distributions from their dependence, making them a suitable fit for this problem \cite{Nelsen2006}. A copula is a multivariate distribution with uniform marginals on $[0,1]$ that connects single-variable models into a joint one \cite{Sklar1959}. 
By choosing different copula types, we can model symmetric, lower-tail, upper-tail, or more complex relationships without restricting the form of the individual risk scores. In biomedical statistics, copula-based methods have been used to model dependence structures in bivariate survival data and multivariate event processes \cite{shih1995copula}. More broadly, related developments in survival analysis have addressed complex data settings such as interval-censored outcomes, underscoring the need for flexible statistical frameworks to model dependence structures and incomplete observations \cite{satten1996interval}. More recently, copula-based approaches have been applied to modeling the spread of breast cancer \cite{gasparini2022natural}.
% By choosing different copula types, we can model symmetric, lower-tail, upper-tail, or more complex relationships without limiting the form of the individual risk scores. In biomedical statistics, copulas have been used for bivariate survival data and multivariate event processes \cite{shih1995copula,satten1996interval}, and more recently for modelling the spread of breast cancer \cite{gasparini2022natural}.
% Pair-copula and vine copula methods have also enhanced risk profiling before surgery, enabling the better identification of low-risk patients compared to standard methods \cite{sahin2025probabilistic}.
Pair-copula and vine copula methods have also been developed for classification tasks, with recent work showing that copula-based classifiers can improve performance when predictors exhibit complex or uneven dependence structures \cite{sahin2025probabilistic}.
% Outside medicine, copula-based fusion has been used to combine different scores in speaker verification and other pattern-recognition tasks, where modelling the dependence between outputs improves results over simple linear methods \cite{cumani2025copula}. 
Outside medicine, copula-based fusion has been used to combine scores in biometric and pattern-recognition tasks, where modeling the dependence between outputs can improve performance compared to simple linear fusion methods \cite{cumani2025copula}.
% Copula concepts have also been included in Bayesian networks for genomic data and in hybrid ML frameworks for survival analysis \cite{zhang2017mixture,kim2025integrating}.
Copula concepts have also been incorporated into Bayesian networks for genomic data and into hybrid machine learning frameworks for modeling dependence in time-to-event and related predictive settings \cite{zhang2017mixture, kim2025integrating}.
To our knowledge, there are no prior published studies that directly model the joint distribution of machine learning--derived clinical and gene-expression risk scores using copulas for breast cancer risk stratification. Most current methods treat the clinical and gene-expression components as either separate models, whose results are averaged, or as features in a single combined model. These approaches do not focus on the dependence between the risk scores and are not designed to detect extreme-risk interactions or mismatched risk patterns.

In this study, we investigate whether combining clinical and gene-expression ML risk scores using copulas can provide useful prognostic structure for breast cancer risk stratification. Using the METABRIC cohort \cite{curtis2012,pereira2016}, we first build supervised ML models separately for clinical variables and high-dimensional mRNA expression, obtaining out-of-fold risk predictions for each patient. These predictions are then converted to values on the unit interval and used to fit Gaussian, Clayton, Gumbel, and Frank copulas, which represent different types of dependence. We assess each model's fit using a bootstrap-calibrated Cram\'{e}r--von Mises test with parameter re-estimation and use the best copula to define both a continuous fused risk score and joint risk groups. We compare survival differences across groups defined by clinical-only, gene-expression-only, and copula-fused risk scores, and supplement the Kaplan--Meier analysis with a competing-risks framework that accounts for non-cancer mortality. To examine transportability, we conducted an independent external evaluation in TCGA under a reduced harmonized specification.
Because the cohorts differed in endpoint, predictor availability, gene coverage, and expression platform, this analysis evaluates the
transferability of the score-fusion procedure rather than directly validating the richer primary METABRIC model.

The primary aims of this study are:
\begin{enumerate}
    \item To develop clinical and gene-expression machine learning risk scores for predicting five-year cancer-specific mortality in the METABRIC cohort.
    \item To model the joint dependence structure of these risk scores using parametric copulas from multiple families.
    \item To evaluate whether median-defined joint clinical and gene-expression score groups show statistically distinct long-term outcome patterns.
    \item To identify subgroups characterized by concordant or discordant clinical--gene-expression risk and to assess their survival outcomes.
    \item To externally evaluate the transportability of the copula-based fusion procedure in TCGA under a reduced harmonized predictor and
endpoint specification.
\end{enumerate}
Throughout, we frame this study as methodological and exploratory: our goal is
to develop and characterize an interpretable, dependence-aware fusion framework,
not to establish a deployment-ready clinical tool.

Our results show that copula-based modeling captures the dependence between clinical and gene-expression ML risk scores in the METABRIC cohort, and that joint risk strata defined by the fused scores identify patient subgroups with clearly different long-term outcomes.  In the independent TCGA evaluation, the copula-fused score showed discrimination comparable to the individual and simple-fusion scores
, with overlapping confidence intervals. The result supports the feasibility of applying the reduced procedure across cohorts but does
not establish superiority or direct validation of the primary model.

The next section reviews related work on clinical and gene-expression prognostic modeling and the application of copulas in biomedical statistics.

%%%%%%%%%%%%%%%%%%%%%%%%%%%%%%%%%%%%%%%%%%%%%%%%%%%%%%%%%%%%

\section*{Related Work}
\label{sec:related}

In this section, we review prior work on clinical and gene-expression prognostic modeling, multimodal machine learning, and copula-based methods, and we position our copula-based fusion of clinical and gene-expression risk scores within this landscape.

\subsection*{Clinical and Gene-expression Prognostic Models in Breast Cancer}

Traditionally, breast cancer prognosis has relied on clinical and pathological factors like tumor size, nodal status, grade, and hormone receptor status. Tools such as the Nottingham Prognostic Index, Adjuvant! Online, and PREDICT are still widely used because they use affordable, routinely collected data and provide clear risk estimates at the bedside \cite{galea1992,ravdin2001,wishart2010}. However, tumors with similar clinical profiles can have very different outcomes, showing the biological diversity of breast cancer and the limitations of using only clinical models.

High-throughput molecular profiling has led to more biologically informed methods for assessing risk. Tests like Oncotype DX, MammaPrint, and PAM50 use gene-expression patterns to generate recurrence scores that help guide chemotherapy decisions for early-stage breast cancer \cite{paik2004,van2002,parker2009}. These tests reveal tumor biology that standard pathology cannot capture and are now part of clinical decision-making in selected settings. Still, studies show that gene-expression tests often derive much of their predictive strength from proliferation-related genes and may only modestly improve risk prediction when added to strong clinical models. Because of this, researchers increasingly view clinical and gene expression data as complementary sources of risk.

The METABRIC cohort is one of the most widely studied datasets combining clinical and gene-expression data in breast cancer. Neapolitan and Jiang used METABRIC to show that integrating clinical variables with genome-wide expression data improved survival prediction compared with using clinical data alone, and they also found that random survival forests performed well in this heterogeneous setting \cite{neapolitan2015integrated}. Liu and colleagues examined differences across molecular subtypes and found that the prognostic contributions of clinical and gene-expression variables vary across subtypes, with gene-expression features being more informative in some groups and clinical variables remaining useful in others \cite{liu2014breast}.
% Liu and colleagues examined differences across molecular subtypes and found that clinical variables often retained substantial prognostic value even when gene-expression data were available \cite{liu2014breast}. 
These findings support the view that clinical and gene-expression features capture related but non-identical aspects of risk, motivating two-view approaches rather than simply collapsing all variables into a single feature set.

\subsection*{Multimodal Machine Learning and Current Fusion Strategies}

Researchers have used machine learning (ML) methods on clinical, gene-expression, and multi-omics data to identify complex patterns and interactions. Techniques such as random forests, gradient boosting, penalized regression, and deep neural networks have all been used to derive prognostic scores from molecular and clinical data, often outperforming traditional regression models in research settings. However, ML models that rely heavily on molecular data can be sensitive to batch effects, cohort shifts, and platform-specific differences, which may reduce their stability in new datasets.

Many studies now focus on combining clinical and molecular information in unified ML frameworks. Most often, this is done by merging clinical and molecular features into a single large predictor set or by combining separate risk scores with fixed or learned weights. For example, Neapolitan and Jiang’s METABRIC analysis combined clinical and gene-expression features in a single random survival forest model \cite{neapolitan2015integrated}. Stacking and ensemble methods may capture nonlinearities and interactions, but they generally do not estimate a separate parametric
dependence model for the view-specific risk scores. The copula formulation, therefore, offers a different descriptive representation of
cross-view dependence rather than necessarily a more flexible predictive model.

\subsection*{Copulas in Biomedical Statistics and Model Fusion}

Copula models provide a structured approach to describing the relationship between multiple variables while separating their marginal behavior from their dependence structure \cite{Nelsen2006}. In biostatistics, copulas have been used to analyze bivariate and multivariate survival outcomes, allowing flexible modeling even when data are censored. Shih and Louis developed methods for estimating association in copula models for bivariate survival data, showing that these approaches can accommodate complex relationships in time-to-event outcomes \cite{shih1995copula}. More recently, Gasparini and Humphreys used a copula-based model to link tumor growth, nodal spread, and distant metastasis in breast cancer while preserving interpretability of the component relationships \cite{gasparini2022natural}.

Copulas have also been used to model genomic and biomedical data outside survival analysis. Zhang and Shi developed a mixture-copula Bayesian network for multimodal genomic data, allowing flexible non-Gaussian dependence structures in network models and improving predictive performance relative to Gaussian Bayesian networks \cite{zhang2017mixture}. Copula-based constructions, including pair-copula and vine copula approaches, have been applied to classification problems, demonstrating improved prediction and calibration when underlying predictor relationships are non-linear or exhibit varying dependence structures \cite{sahin2025probabilistic}.
% In perioperative and other clinical risk settings, pair-copula and vine-copula models have been used for patient risk profiling, and recent work suggests that copula-based classifiers can improve calibration and Brier-score performance when predictors show uneven dependence in the tails \cite{sahin2025probabilistic}.

Outside of medicine, copula-based fusion has been used to combine outputs from multiple machine learning systems. For instance, copula-based fusion techniques have been applied in biometric and pattern recognition contexts, where modeling the joint behavior of system outputs can enhance calibration relative to linear fusion methods \cite{cumani2025copula}.
% For example, copula-based score fusion has been used in speaker verification, where modelling the joint distribution of system scores can improve calibration relative to linear fusion \cite{cumani2025copula}. 
% Copula ideas have also been incorporated into broader hybrid ML frameworks for survival and treatment-effect modelling \cite{kim2025integrating}. 
Copula-based approaches have also been integrated into hybrid machine learning frameworks to capture dependence structures in time-to-event and other predictive settings \cite{kim2025integrating}.
Taken together, these studies suggest that copulas can serve as a useful tool for combining predictive models while accounting for dependence between their outputs.

\subsection*{Positioning of the Present Study}

Despite this growing literature, the use of copulas to combine machine-learning risk scores from separate clinical and gene-expression sources remains largely unexplored in oncology. Most breast cancer prognosis studies either combine clinical and molecular variables into one feature set or use linear or additive strategies to combine model outputs, thereby assuming relatively simple relationships between views. Previous copula work in biomedicine has focused primarily on survival outcomes, metastasis, or biomarker dependence \cite{shih1995copula,gasparini2022natural,zhang2017mixture}, rather than on fusing machine-learning risk scores.

This study addresses that gap by using parametric copulas, including Gaussian, Clayton, Gumbel, and Frank families, to analyze the empirical ranks of out-of-fold clinical and gene-expression machine learning risk scores in the METABRIC cohort. By modeling different types of dependence, including symmetric dependence and lower- or upper-tail emphasis, we examine how clinical and gene-expression risks co-vary across the risk spectrum. This copula-based approach supports both a fused continuous score and descriptively interpretable joint score groups, allowing us to examine concordant and discordant risk patterns. To our knowledge, this is among the first applications of copula-based fusion to combine multimodal machine-learning risk scores for breast cancer prognosis, and it provides a general framework for integrating distinct predictive views in biomedical risk modeling.

In the next section, we outline the probabilistic and statistical tools underlying our copula-based approach to combining clinical and gene-expression machine learning risk scores.

%%%%%%%%%%%%%%%%%%%%%%%%%%%%%%%%%%%%%%%%%%%%%%%%%%%%%%
\section*{Preliminaries}
\label{sec:preliminaries}

In this section, we summarize the main probabilistic and statistical tools used in our analysis. These include copulas and tail dependence, the specific copula families we considered (Gaussian, Clayton, Gumbel, and Frank), the Cram\'er--von Mises goodness-of-fit test with parametric bootstrap, the machine learning models used to build risk scores, and the survival analysis framework for evaluation.

\subsection*{Copulas and Pseudo-observations}

A bivariate copula $C:[0,1]^2 \to [0,1]$ is a joint distribution function with uniform marginals that captures the dependence structure between two random variables independently of their marginal scales \cite{Sklar1959,Nelsen2006,Joe1997}. By Sklar's theorem \cite{Sklar1959}, any bivariate cumulative distribution function $H$ with continuous marginals $F_X$ and $F_Y$ can be uniquely decomposed as
\begin{equation}
H(x,y) \;=\; C\big(F_X(x), F_Y(y)\big),
\end{equation}
where $C$ is the copula of $(X,Y)$.

In this work, $X$ and $Y$ are the clinical and gene-expression machine learning risk scores, respectively. To work on the copula scale, we convert these scores into \emph{pseudo-observations} by ranking:
\begin{equation}
U_i \;=\; \frac{\mathrm{rank}(X_i)}{n+1},
\qquad
V_i \;=\; \frac{\mathrm{rank}(Y_i)}{n+1},
\qquad i=1,\dots,n,
\end{equation}
% where $n$ is the number of patients and $\mathrm{rank}(\cdot)$ is the average rank.

where $n$ denotes the number of patients, and $\operatorname{rank}(X_i)$ and
$\operatorname{rank}(Y_i)$ denote the midranks of $X_i$ and $Y_i$,
respectively, among the clinical and gene-expression risk scores. The pairs $(U_i,V_i)$ lie in $(0,1)^2$ and are approximately sampled from the underlying copula. Any subsequent quantity computed from $C$ therefore depends only on the joint behaviour of the risk scores and not on their marginal calibration.

\subsection*{Tail Dependence}

Tail dependence measures the strength of association in the extremes of a bivariate distribution. For a copula $(U,V)$ with continuous marginals, the \emph{upper} and \emph{lower} tail-dependence coefficients are defined as \cite{Nelsen2006,Joe1997}
\begin{align}
\lambda_U &= \lim_{q \to 1^-} \Pr(V > q \mid U > q), \\
\lambda_L &= \lim_{q \to 0^+} \Pr(V \le q \mid U \le q),
\end{align}
whenever the limits exist. Values of $\lambda_U$ or $\lambda_L$ close to $1$ indicate strong clustering of extreme events, whereas values near $0$ indicate tail independence.

In our setting, $(U,V)$ represents the ranks of clinical and gene-expression risk scores. A positive upper-tail coefficient $\lambda_U$ indicates that patients who are extremely high-risk clinically also tend to be extremely high-risk on the gene-expression score, while a positive lower-tail coefficient $\lambda_L$ indicates co-occurrence of very low predicted risks. These coefficients provide an interpretable summary of how the two modalities interact in the most clinically relevant regions of the risk spectrum.

\subsection*{Gaussian, Clayton, Gumbel, and Frank Copulas}

We focus on four parametric copula families that capture different dependence patterns: a symmetric elliptical copula (Gaussian), two Archimedean copulas with asymmetric tail behavior (Clayton and Gumbel), and an Archimedean copula with symmetric dependence but no asymptotic tail dependence (Frank) \cite{Nelsen2006,Joe1997}.

\paragraph{Gaussian copula.}
Let $\Phi$ denote the standard normal cdf and $\Phi_\rho$ the bivariate normal cdf with correlation $\rho \in (-1,1)$. The Gaussian copula is
\begin{equation}
C_\rho^{\text{Gauss}}(u,v)
= \Phi_\rho\!\big(\Phi^{-1}(u), \Phi^{-1}(v)\big),
\qquad (u,v)\in(0,1)^2.
\end{equation}
We estimate $\rho$ from Kendall's rank correlation $\tau$ via
\begin{equation}
\rho \;=\; \sin\!\left(\frac{\pi}{2}\tau\right),
\end{equation}
which follows from the elliptical structure \cite{Nelsen2006}. For $|\rho|<1$, the Gaussian copula is tail-independent, with $\lambda_L=\lambda_U=0$ \cite{Nelsen2006}.

\paragraph{Clayton copula.}
The Clayton copula \cite{Clayton1978} is an Archimedean copula with parameter $\theta>0$ and cdf
\begin{equation}
C_\theta^{\text{Clay}}(u,v)
= \big(u^{-\theta}+v^{-\theta}-1\big)^{-1/\theta},
\qquad (u,v)\in(0,1)^2.
\end{equation}
Clayton exhibits lower-tail but not upper-tail dependence:
\begin{equation}
\lambda_L(\theta)=2^{-1/\theta},
\qquad
\lambda_U(\theta)=0.
\end{equation}
Its Kendall's tau satisfies
\begin{equation}
\tau(\theta)=\frac{\theta}{\theta+2},
\qquad\Longleftrightarrow\qquad
\theta=\frac{2\tau}{1-\tau},
\end{equation}
which we use to estimate $\theta$ from the empirical $\hat\tau(U,V)$.

\paragraph{Gumbel copula.}
The Gumbel copula \cite{Gumbel1960,Hougaard1986} is an Archimedean copula with parameter $\theta\ge1$ and cdf
\begin{equation}
C_\theta^{\text{Gum}}(u,v)
=
\exp\!\left\{
-\Big[(-\log u)^\theta+(-\log v)^\theta\Big]^{1/\theta}
\right\},
\qquad (u,v)\in(0,1)^2.
\end{equation}
Gumbel has positive upper-tail dependence but no lower-tail dependence:
\begin{equation}
\lambda_U(\theta)=2-2^{1/\theta},
\qquad
\lambda_L(\theta)=0.
\end{equation}
Kendall's tau is
\begin{equation}
\tau(\theta)=1-\frac{1}{\theta},
\qquad\Longleftrightarrow\qquad
\theta=\frac{1}{1-\tau}, \quad \tau\in(0,1).
\end{equation}
Thus $\theta$ can again be estimated from $\hat\tau(U,V)$, and $\lambda_U$ then follows in closed form.

\paragraph{Frank copula.}
The Frank copula is an Archimedean copula that allows symmetric positive or negative dependence without asymptotic tail dependence \cite{Nelsen2006,Joe1997}. Its cdf is
\begin{equation}
C_\theta^{\text{Frank}}(u,v)
=
-\frac{1}{\theta}
\log\!\left(
1+\frac{(\mathrm{e}^{-\theta u}-1)(\mathrm{e}^{-\theta v}-1)}{\mathrm{e}^{-\theta}-1}
\right),
\qquad \theta\neq0,
\end{equation}
with independence recovered in the limit as $\theta\to0$. Like the Gaussian copula, the Frank copula has
\begin{equation}
\lambda_L(\theta)=\lambda_U(\theta)=0.
\end{equation}
Its parameter can be related to Kendall's tau through the Debye function, so in practice $\theta$ is obtained numerically from the empirical $\hat\tau(U,V)$.

Together, these four families allow us to compare symmetric dependence without tail concentration (Gaussian and Frank) against asymmetric models that emphasize lower-tail or upper-tail agreement (Clayton and Gumbel).

These four families were chosen as a compact, standard set spanning the
qualitatively distinct dependence structures most relevant to risk-score fusion:
symmetric dependence without tail concentration (Gaussian and Frank), lower-tail
dependence (Clayton), and upper-tail dependence (Gumbel). Other families, such as
the Joe and two-parameter BB copulas, rotated Archimedean copulas, and vine
constructions for higher-dimensional settings, may better capture particular
biomedical score distributions and represent a natural direction for future work.
In the present study, family choice is decided empirically through formal
goodness-of-fit testing rather than fixed a priori.

\subsection*{Cram\'er--von Mises Goodness-of-fit for Copulas}

To assess how well a parametric copula family fits the empirical dependence between clinical and gene-expression risk scores, we use a Cram\'er--von Mises (CvM) statistic based on the empirical copula \cite{GenestRivest1993,GenestRemillard2008,Genest2007}. Let
\begin{equation}
C_n(u,v)
=
\frac{1}{n}\sum_{i=1}^n
\mathbf{1}\{U_i\le u,\;V_i\le v\}
\end{equation}
denote the empirical copula of $(U_i,V_i)_{i=1}^n$, and let $C_{\hat\theta}$ be a fitted parametric copula with parameter(s) $\hat\theta$. The CvM discrepancy is
\begin{equation}
S_n
=
\frac{1}{n}\sum_{i=1}^n
\big[C_n(U_i,V_i)-C_{\hat\theta}(U_i,V_i)\big]^2.
\end{equation}

Because the null distribution of $S_n$ depends on both the copula family and the fitted parameter, we approximate its sampling distribution by a parametric bootstrap \cite{GenestRemillard2008,Genest2007}. Specifically, we repeatedly simulate pseudo-observations $(U_i^{(b)},V_i^{(b)})_{i=1}^n$ from the fitted copula $C_{\hat\theta}$. For each bootstrap sample, we re-estimate the copula parameter, recompute the corresponding CvM statistic $S_n^{(b)}$, and estimate the $p$-value as
\begin{equation}
\hat p
=
\frac{1+\sum_{b=1}^B \mathbf{1}\{S_n^{(b)}\ge S_n\}}{B+1},
\end{equation}
with $B=1000$ replications in our main METABRIC analysis. 
Among the prespecified candidate families, the family with the smallest observed Cram\'er--von Mises discrepancy was selected as the primary descriptive fit. Bootstrap $p$-values were used only to assess evidence of inadequate fit. Because the same score sample was
used for family comparison and description, the selected family should not be interpreted as a prospectively validated optimum.

% Among candidate families, the best fit is chosen as the one with the smallest Cram\'er--von Mises statistic $S_n$; the bootstrap $p$-value is retained only as an adequacy diagnostic for the selected family, because a $p$-value is not a comparable model-selection score across families.

\subsection*{Machine Learning Risk Scores}

We construct separate machine-learning risk scores from the clinical and gene-expression views using standard classifiers on tabular data. Let $Z^{(\mathrm{clin})}$ denote the matrix of clinical predictors and $Z^{(\mathrm{expr})}$ the matrix of gene-expression predictors, and let $Y$ be the binary indicator of 5-year cancer-specific mortality.

For each view, we consider the following supervised learning models:
\begin{itemize}
\item \textbf{Logistic regression with elastic-net penalty (LR).} A generalized linear model that maps features to log-odds of the outcome, with combined $\ell_1$ and $\ell_2$ penalties to encourage sparsity and stabilize estimates \cite{Cox1958,ZouHastie2005}.
\item \textbf{Random forest (RF).} An ensemble of decision trees trained on bootstrap samples with feature subsampling at each split, which captures nonlinearities and interactions while reducing variance through averaging \cite{Breiman2001}.
\item \textbf{Gradient boosting (GB).} An additive ensemble of shallow trees fitted sequentially, where each new tree is trained to reduce the residual error of the current ensemble \cite{Friedman2001}.
\item \textbf{Extreme gradient boosting (XGB).} A regularized implementation of gradient boosting with system-level optimizations and additional shrinkage and penalty terms for improved performance on tabular data \cite{Chen2016}, used when the \texttt{xgboost} library is available.
\end{itemize}

For each model and each view, we perform 5-fold stratified cross-validation and use out-of-fold predicted probabilities $\hat p_i^{(\mathrm{clin})}$ and $\hat p_i^{(\mathrm{expr})}$ as clinical and gene-expression risk scores, respectively. This cross-validation scheme avoids optimistic bias by ensuring that each patient's risk score is obtained from a model that did not use that patient's outcome for training. The final risk scores used for copula fitting correspond to the best-performing model in each view, as selected by cross-validated ROC-AUC.

\subsection*{Survival Analysis and ROC-AUC}

Although the primary endpoint for model training is 5-year cancer-specific mortality, the METABRIC cohort also includes detailed follow-up times and censoring information. To visualize and compare long-term outcome patterns across joint risk strata, we use the Kaplan--Meier estimator \cite{KaplanMeier1958}. For a given subgroup, let $t_1<t_2<\dots<t_J$ be the distinct event times, and let $d_j$ and $r_j$ denote the numbers of events and patients at risk at time $t_j$. The Kaplan--Meier estimate of the survival function is
\begin{equation}
\hat S(t)=\prod_{t_j\le t}\left(1-\frac{d_j}{r_j}\right),
\end{equation}
which provides a non-parametric estimate of the probability of remaining event-free beyond time $t$ while accommodating right-censoring.

For discrimination performance, we use the area under the receiver operating characteristic curve (ROC-AUC) as the primary metric. Given model scores $s_i$ and binary outcomes $Y_i$, ROC-AUC is the probability that a randomly chosen event case receives a higher score than a randomly chosen non-event case,
\begin{equation}
\mathrm{AUC}
=
\Pr(s^+>s^-)
+\frac{1}{2}\Pr(s^+=s^-),
\end{equation}
and summarizes threshold-free ranking performance. ROC-AUC is computed separately for clinical-only, gene-expression-only, and copula-fused risk scores.

In the next section, we describe the METABRIC and TCGA datasets used in the primary analysis and external evaluation.

%%%%%%%%%%%%%%%%%%%%%%%%%%%%%%%%%%%%%%%%%%%%%%%%%%%%%%%
\section*{Dataset Description}
\label{sec:dataset}

In this study, we used two breast cancer datasets. The primary development dataset was the Molecular Taxonomy of Breast Cancer International Consortium (METABRIC), a large Canada--UK project that profiled primary breast tumors with clinical, pathological, and molecular measurements. METABRIC includes nearly 2,000 patients with long-term follow-up and is widely used as a benchmark dataset for breast cancer subtype analysis and prognosis. For external evaluation, we used an independent breast cancer cohort from The Cancer Genome Atlas (TCGA), constructed to provide harmonized clinical and gene-expression data for out-of-cohort testing.

\subsection*{METABRIC Dataset}

For the primary analyses, we used the processed file \texttt{METABRIC\_RNA\_Mutation.csv}. This file combines curated clinical attributes, mRNA expression $z$-scores for hundreds of genes, mutation indicators, and survival outcomes for each patient. The underlying data originate from the METABRIC project, first reported by Curtis et al.~\cite{curtis2012} and subsequently expanded by Pereira et al.~\cite{pereira2016}; we accessed the publicly available, curated, pre-processed CSV file distributed via Kaggle~\cite{metabricKaggle}. Using this processed file provided a transparent and reproducible starting point for the METABRIC-based analyses, while the original studies of Curtis et al.~\cite{curtis2012} and Pereira et al.~\cite{pereira2016} remain the primary data source and should be cited as such.

\subsubsection*{METABRIC Clinical Variables}

The clinical portion of the METABRIC file contains one row per patient and includes baseline and tumor-related variables. These include age at diagnosis, type of breast surgery, cancer type, detailed histological subtype, tumor cellularity, histological grade, oestrogen receptor, progesterone receptor, and HER2 status, molecular subtype labels such as PAM50 and three-gene classifier subtype, treatment indicators, inferred menopausal state, integrative cluster assignment, tumor laterality, lymph-node information, mutation count, Nottingham Prognostic Index, tumor size, and tumor stage. Some variables contain missing entries because of incomplete source records.

The file also provides survival information, including \texttt{overall\_survival\_months}, \texttt{overall\_survival}, and \texttt{death\_from\_cancer}. These variables were used to define the primary 5-year cancer-specific death endpoint in the METABRIC analysis.

\subsubsection*{METABRIC Gene-expression Variables}

The molecular portion of the METABRIC file contains high-dimensional data, including mRNA expression $z$-scores for 489 genes and binary mutation indicators. In the primary pipeline, the gene-expression view was defined using the available numeric mRNA expression features after excluding clinical variables, identifier variables, survival variables, and mutation indicator columns. Thus, the main molecular view in the analysis is gene expression rather than a combined genomic view.

Taken together, the METABRIC gene-expression and clinical variables provide a natural two-view structure for modeling clinical and molecular risk separately and then studying their joint behavior.

\subsection*{TCGA External Evaluation Dataset}

For external evaluation, we constructed an independent TCGA breast cancer dataset using the cBioPortal study \textit{Breast Invasive Carcinoma (TCGA, PanCancer Atlas)}. From this study, we downloaded the clinical patient data and the mRNA expression $z$-scores relative to all samples (RNA Seq V2 RSEM / log RNA Seq V2 RSEM), and then used a preprocessing pipeline in R to clean, harmonize, and merge these files into a final evaluation dataset.

The TCGA clinical data included patient identifier, age at diagnosis, tumor stage, tumor size, nodal stage, metastasis stage, subtype, race, radiation therapy, overall survival time, and overall survival status. These variables were cleaned and standardized, and tumor stage was harmonized into broad stage categories. The gene-expression component consisted of gene-level mRNA expression $z$-scores. After cleaning the TCGA expression matrix and harmonizing gene names, overlapping genes were identified by intersecting the available TCGA gene expression variables with the METABRIC gene expression set.

Some TCGA-specific clinical variables were retained in the derived dataset for descriptive and harmonization purposes, but variables not shared with METABRIC were excluded from the final harmonized predictor set used for external evaluation.

\subsection*{Working Datasets for Analysis}

We used these two datasets in two related but distinct ways.

First, for the primary METABRIC analysis, we used \texttt{METABRIC\_RNA\_Mutation.csv} to define a 5-year cancer-specific death endpoint, construct clinical and gene-expression predictor views, train machine learning models, generate out-of-fold risk scores, fit copula models, and study joint risk stratification in the METABRIC cohort.

Second, for external evaluation, we created harmonized METABRIC and TCGA datasets by retaining only predictors available in both cohorts. This yielded a shared clinical view and a shared gene-expression view with 476 common gene-expression features. In this harmonized setting, we defined a consistent 5-year overall survival endpoint in both datasets, trained the clinical and gene-expression models in METABRIC, and then evaluated the resulting clinical, gene-expression, and copula-fused scores in the independent TCGA cohort.

The next section explains the modeling methodology in detail, including endpoint construction, predictor-view definition, machine learning pipelines, copula-based dependence modeling, and the external evaluation procedure.

%%%%%%%%%%%%%%%%%%%%%%%%%%%%%%%%%%%%%%%%%%%%%%%%%%%%%%%%%%%%%
\section*{Methodology}
\label{sec:methodology}

\subsection*{Overview and Guiding Question}

In this section, we describe the development, internal, and external evaluations of clinical and gene-expression machine learning (ML) risk scores for breast cancer prognosis. Our primary question is: for the same patients, how do risk scores derived from routine clinical data and high-dimensional gene-expression data compare when predicting 5-year mortality, and what does their joint behavior reveal about patients at particularly high risk?

We approached the problem as a two-view prediction task. In the METABRIC cohort, we first created a binary 5-year cancer-specific death endpoint and retained only patients with sufficient information to determine this outcome. We then divided the predictors into two views. The clinical view included routinely collected demographic, tumor, receptor, subtype, and treatment-related variables. The gene-expression view included mRNA expression features from the same patients.

For each view, we trained several supervised ML models and obtained 5-fold out-of-fold predicted probabilities for every patient. These out-of-fold probabilities defined the clinical and gene-expression risk scores. We then converted the two scores to pseudo-observations on the unit square and fitted several bivariate copula families to model their dependence structure. Finally, we assessed whether the resulting joint risk patterns were associated with long-term outcome differences using Kaplan--Meier and competing-risks analyses.

To address external evaluation, we then harmonized METABRIC and TCGA by retaining only the clinical and gene-expression predictors available in both cohorts and by defining a consistent 5-year overall survival endpoint in both datasets. Models were trained in METABRIC and evaluated without retraining in the independent TCGA cohort.

\subsection*{METABRIC endpoint definition and analytic cohort}

All internal model development used the processed METABRIC dataset. For each patient, the file provides overall survival time in months, an overall survival status, and a specific indicator for death from breast cancer. In the code, these variables are \texttt{overall\_survival\_months}, \texttt{overall\_survival}, and \texttt{death\_from\_cancer}.

For the primary METABRIC analysis, we defined a binary 5-year cancer-specific death endpoint using a fixed 60-month window. Let
\[
T = \texttt{overall\_survival\_months}.
\]
When available, \texttt{death\_from\_cancer} was used as the main source of event status. Patients recorded as ``Died of Disease'' were treated as cancer-specific events, patients recorded as ``Living'' were treated as non-events, and patients recorded as ``Died of Other Causes'' were treated as competing events and excluded from the primary 5-year cancer-specific endpoint analysis. If the cancer-specific field was unavailable, \texttt{overall\_survival} was used as fallback status information.

The 5-year endpoint \(Y\) was defined as
\[
Y =
\begin{cases}
1, & \text{if cancer-specific death occurred within 60 months,}\\[2pt]
0, & \text{if the patient survived beyond 60 months or died from cancer after 60 months,}\\[2pt]
\text{missing}, & \text{if follow-up was insufficient for 5-year determination or if death was from another cause.}
\end{cases}
\]

Patients with missing \(Y\) were excluded from the primary endpoint analysis. This approach avoids assigning 5-year status to patients without enough follow-up and separates cancer-specific death from non-cancer competing deaths in the main METABRIC endpoint.

\subsection*{Clinical and Gene-expression Predictor Views}

For the METABRIC analysis, we separated predictors into clinical and gene-expression views.

The clinical view was built from the clinical and pathological variables listed in the processed METABRIC file, including age at diagnosis, type of breast surgery, cancer type and histology, tumor cellularity and grade, oestrogen, progesterone, and HER2 receptor status, subtype labels, treatment indicators, menopausal state, integrative cluster assignment, laterality, lymph-node information, mutation count, Nottingham prognostic index, tumor size, and tumor stage. Survival-related variables were removed from the predictor set, and the patient identifier was excluded from modeling.

The proportion of missing values among the clinical predictors used for
modeling is reported in Table~\ref{tab:missingness}. Within the analytic cohort
($n=1{,}363$), missingness was confined to a minority of variables, the
largest being tumor stage ($23.99\%$), three-gene classifier subtype
($10.71\%$), and primary tumor laterality ($5.43\%$); all other clinical
predictors had under $4\%$ missingness, and fifteen predictors had none.
Numeric variables were median-imputed, and categorical variables were imputed
using the most frequent category, with all imputation performed inside the
cross-validation training folds so that no held-out-fold information was used.

\begin{table}[ht]
\centering
\caption{Proportion of missing values for clinical predictors in the
METABRIC analytic cohort ($n=1{,}363$). Predictors not listed had no missing
values.}
\label{tab:missingness}
\begin{tabular}{lc}
\hline
Clinical predictor & Missing (\%) \\
\hline
Tumor stage & 23.99 \\
Three-gene classifier subtype & 10.71 \\
Primary tumor laterality & 5.43 \\
Cellularity & 3.01 \\
Mutation count & 2.64 \\
Neoplasm histologic grade & 2.49 \\
ER status (by IHC) & 1.32 \\
Type of breast surgery & 0.73 \\
Tumor size & 0.73 \\
Tumor other histologic subtype & 0.59 \\
Cancer type detailed & 0.59 \\
Oncotree code & 0.59 \\
\hline
\end{tabular}
\end{table}

The gene-expression view was defined as the set of numeric mRNA expression features after removing clinical variables, identifier variables, survival variables, and mutation indicator columns. In the pipeline, all available gene-expression features of this type were retained rather than restricting the analysis to a variance-selected subset. We also conducted a sensitivity analysis in which the gene-expression model was re-evaluated with varying numbers of retained features.

\subsection*{Supervised Machine Learning for METABRIC Risk Scores}

We trained supervised classification models separately on the clinical and gene-expression views to predict the binary 5-year cancer-specific death endpoint. Let \(X_{\text{clin}}\) denote the clinical predictor matrix, \(X_{\text{expr}}\) the gene-expression predictor matrix, and \(Y\) the binary outcome.

For the clinical view, predictors were separated into numeric and categorical variables. Numeric variables were imputed using the median and standardized to mean zero and variance one. Categorical variables were imputed using the most frequent category and then one-hot encoded. For the gene-expression view, numeric features were median-imputed and standardized. All preprocessing steps were embedded in the cross-validation pipeline, so that imputation, scaling, and encoding were fit only on the training folds and then applied to the held-out folds.

For clarity and to make the absence of information leakage verifiable by
inspection, the procedure for each view can be stated step by step:
(i) the analytic cohort is partitioned into five stratified folds;
(ii) for each fold, all preprocessing transformers (median or most-frequent
imputation, standardization, and one-hot encoding) are fitted on the four
training folds only;
(iii) the fitted transformers are applied to the held-out fold to produce its
feature matrix;
(iv) the classifier is trained on the transformed training folds and used to
predict probabilities for the held-out fold;
(v) steps (ii)--(iv) are repeated so that every patient receives an out-of-fold
predicted probability obtained without using that patient's outcome for fitting.
The resulting out-of-fold probabilities are used as the view-specific risk
scores. Because preprocessing and model fitting occur entirely within the
training folds, no held-out-fold information can influence the risk scores.

We evaluated four candidate classifiers in each view: elastic-net logistic regression, random forest, gradient boosting, and XGBoost when available. We used 5-fold stratified cross-validation and obtained out-of-fold predicted probabilities for every patient. These out-of-fold probabilities were used as the view-specific ML risk scores:
\[
\hat{p}_{\text{clin}} = (\hat{p}_{\text{clin},1},\dots,\hat{p}_{\text{clin},n}), \qquad
\hat{p}_{\text{expr}} = (\hat{p}_{\text{expr},1},\dots,\hat{p}_{\text{expr},n}).
\]

For each view, the model with the highest cross-validated ROC-AUC was selected as the main risk-score generator. We also computed bootstrap confidence intervals for the resulting ROC-AUC values. In addition, for the gene-expression view, we evaluated sensitivity to feature count by repeating the random forest analysis using several retained feature sizes.

\subsection*{Copula Modeling of Joint Risk Scores}

After obtaining out-of-fold clinical and gene-expression risk scores, we modeled their joint behavior using bivariate copulas. Because copulas are defined on uniform margins, the scores were converted to pseudo-observations by ranking:
\[
U_i = \frac{\mathrm{rank}(\hat{p}_{\text{clin},i})}{n+1}, \qquad
V_i = \frac{\mathrm{rank}(\hat{p}_{\text{expr},i})}{n+1}, \qquad i=1,\dots,n.
\]

We considered four parametric copula families: Gaussian, Clayton, Gumbel, and Frank. Parameters were estimated from Kendall's \(\tau\). For the Gaussian copula,
\[
\hat{\rho} = \sin\!\left(\frac{\pi}{2}\hat{\tau}\right).
\]
For the Clayton copula,
\[
\hat{\theta}_{\text{Clay}} = \frac{2\hat{\tau}}{1-\hat{\tau}},
\]
with a small positive lower bound when needed. For the Gumbel copula,
\[
\hat{\theta}_{\text{Gum}} = \frac{1}{1-\hat{\tau}},
\]
with the constraint \(\hat{\theta}_{\text{Gum}} \ge 1\). For the Frank copula, the parameter was obtained numerically by solving the standard relationship between Kendall's \(\tau\) and the Frank copula parameter.

For each fitted copula, we recorded the corresponding tail-dependence coefficients when applicable. The Gaussian and Frank copulas do not exhibit asymptotic tail dependence, whereas the Clayton and Gumbel copulas capture lower-tail and upper-tail dependence, respectively.

To assess goodness-of-fit, we used a Cram\'er--von Mises statistic based on the empirical copula:
\[
C_n(U_i,V_i)=\frac{1}{n}\sum_{j=1}^n \mathbf{1}\{U_j \le U_i,\;V_j \le V_i\},
\]
and
\[
S=\frac{1}{n}\sum_{i=1}^n \left(C_n(U_i,V_i)-C(U_i,V_i;\hat{\theta})\right)^2.
\]

We calibrated this statistic with a parametric bootstrap. For each copula family, bootstrap samples were simulated from the fitted copula, the copula parameter was re-estimated within each bootstrap sample, and the Cram\'er--von Mises statistic was recomputed. The bootstrap \(p\)-value was estimated as
\[
\hat{p}=\frac{1+\sum_{b=1}^B \mathbf{1}\{S^{(b)} \ge S\}}{B+1}.
\]
The best-fitting dependence model was selected as the copula with the smallest Cram\'er--von Mises statistic, since the goodness-of-fit statistic directly measures the discrepancy between the empirical and fitted copulas; the bootstrap \(p\)-value was retained only as an adequacy diagnostic for the selected family, not as a comparative model-selection score. We also generated empirical heatmaps and fitted heatmaps, along with contour overlays, for visual comparison.

Using the selected copula, we defined a copula-based fused joint risk score by evaluating the copula on pseudo-observations derived from the clinical and gene-expression scores. To assess whether the copula contributed discrimination beyond trivial score combination, we additionally computed three simple score-level fusion baselines --- the rank average of the two pseudo-observations, the average of the two predicted probabilities, and a logistic-regression stacker of the two out-of-fold scores (itself evaluated by 5-fold cross-validation so that no patient contributed to both training and testing of the stacker) --- and compared all scores by ROC-AUC with matched bootstrap confidence intervals.

\subsection*{Joint Risk Strata, Kaplan--Meier, and Competing-risks Analyses}

To relate the joint risk structure to patient outcomes, we defined four joint risk strata using the sample medians of the clinical and gene-expression risk scores. These sample derived median groups were used for descriptive outcome visualization and were not intended as validated clinical risk
categories, treatment thresholds, or prospective decision rules. These groups are given below:
\begin{itemize}
    \item \textbf{low--low}: clinical score \(\le m_{\text{clin}}\) and gene-expression score \(\le m_{\text{expr}}\),
    \item \textbf{high-clinical-only}: clinical score \(> m_{\text{clin}}\) and gene-expression score \(\le m_{\text{expr}}\),
    \item \textbf{high-expression-only}: clinical score \(\le m_{\text{clin}}\) and gene-expression score \(> m_{\text{expr}}\),
    \item \textbf{high-both}: clinical score \(> m_{\text{clin}}\) and gene-expression score \(> m_{\text{expr}}\).
\end{itemize}

We then evaluated long-term outcomes in two complementary ways. First, we estimated Kaplan--Meier curves for overall (all-cause) survival on the full METABRIC cohort ($n=1{,}904$), coding an event as death from any cause (\texttt{overall\_survival}$=0$), and compared groups with log-rank tests. Second, because METABRIC includes deaths from causes other than breast cancer, we also performed a competing-risks analysis in which cancer death was treated as the event of interest and other-cause death as the competing event, again on the full cohort. Both analyses used identical risk strata: the strata cutoffs were the medians of the analytic out-of-fold scores, and patients were assigned out-of-fold scores if they belonged to the analytic cohort or genuine out-of-sample model predictions if they had been excluded from the 5-year endpoint (and hence were never used to train any model). This ensures that the risk-group definitions used for outcome comparison are not obtained in-sample.

\subsection*{External Evaluation in the TCGA Cohort}

To evaluate out-of-cohort generalizability, we performed an external evaluation using an independent TCGA breast cancer cohort after model development in the METABRIC cohort. Because the two cohorts differed in variable structure, we first harmonized variable names, coding, and predictor availability across datasets.

The TCGA evaluation dataset was constructed by combining patient-level clinical information with gene-level mRNA expression $z$-scores. Clinical variables were cleaned and standardized, including age at diagnosis, tumor stage, tumor size, nodal stage, metastasis stage, subtype, race, radiation therapy, overall survival time, and overall survival status. These additional TCGA-specific clinical variables were retained in the derived dataset for descriptive and harmonization purposes, but variables not shared with METABRIC were excluded from the final harmonized predictor set used for external evaluation. Tumor stage values were harmonized into broad stage categories. Gene-expression features were cleaned in parallel, and overlapping genes were identified by intersecting the available TCGA expression variables with the METABRIC gene-expression set, after harmonizing variable names.

For the final external-evaluation pipeline, we retained only predictors available in both cohorts. This yielded a harmonized clinical view comprising the shared clinical variables, and a harmonized gene-expression view comprising the intersection of common gene-expression features, resulting in 476 shared gene-expression features. In the implemented pipeline, the shared clinical variables were age at diagnosis and tumor stage, while the harmonized gene-expression view was defined by the common genes present in both METABRIC and TCGA.

Because METABRIC (microarray) and TCGA (RNA-seq) profiles are generated on different platforms, matching gene names does not guarantee comparable measurement scales. To quantify the resulting distribution shift, we computed, for each shared gene, the standardized mean difference (SMD) between the two cohorts' $z$-scores. The shift was modest for most genes---the median absolute SMD was $0.066$---but not negligible in the tails: $23$ of $476$ genes ($4.8\%$) had an absolute SMD above $0.5$ and $12$ ($2.5\%$) above $1.0$. We report this audit alongside the external results to make the platform mismatch explicit rather than to imply it is removed by name matching.

To ensure that training and testing used the same prediction target, we defined a consistent 5-year overall survival endpoint in both cohorts. Let
\[
T = \texttt{overall\_survival\_months}.
\]
The external-evaluation endpoint was defined as
\[
Y^{\ast} =
\begin{cases}
1, & \text{if death occurred within 60 months,}\\[2pt]
0, & \text{if the patient was alive beyond 60 months or died after 60 months,}\\[2pt]
\text{missing}, & \text{if follow-up was insufficient for 5-year determination.}
\end{cases}
\]
In METABRIC, the processed file encoded \texttt{overall\_survival} as 0 for dead and 1 for alive, whereas in the derived TCGA dataset, the coding was 1 for dead and 0 for alive. The harmonized pipeline handled these dataset-specific encodings explicitly before constructing the common 5-year overall survival endpoint.

Using the harmonized METABRIC dataset, we trained separate clinical and gene-expression machine learning models and obtained out-of-fold predicted probabilities from 5-fold stratified cross-validation. Within METABRIC, these out-of-fold scores were used both to select the best-performing model in each view and to define the training risk-score distributions. After model selection, the final clinical and gene-expression pipelines were fitted on the full harmonized METABRIC training data and then applied unchanged to the independent TCGA cohort to generate external clinical and gene-expression risk scores.

To model dependence between the two METABRIC training scores, we transformed the METABRIC out-of-fold clinical and gene-expression scores to pseudo-observations and fitted candidate Gaussian, Clayton, Gumbel, and Frank copulas. Goodness-of-fit was evaluated using a bootstrap Cram\'er--von Mises procedure, and the best-fitting copula was selected on the METABRIC training scores only. For external fusion, the TCGA clinical and gene-expression scores were mapped onto the METABRIC training-score scale using the empirical distribution of the corresponding METABRIC out-of-fold scores, and the selected copula was then evaluated at these transformed external scores to obtain a fused TCGA risk score. For the external score-level baselines, rank averaging was applied to the METABRIC-mapped pseudo-observations, probability averaging was applied to the two predicted probabilities, and the logistic stacker was fitted on the METABRIC out-of-fold clinical and gene-expression scores and then applied unchanged to TCGA.

Model performance was summarized using ROC-AUC with bootstrap confidence intervals for the clinical-only, gene-expression-only, and copula-fused scores in both the internal METABRIC and external TCGA evaluations. Because model development, copula fitting, and model selection were performed in METABRIC and final evaluation was performed in a completely separate TCGA cohort, this procedure constitutes external evaluation rather than cross-validation. We emphasise, however, that this is an external evaluation of a \emph{harmonized} framework rather than a direct external validation of the primary METABRIC model: to obtain predictors common to both cohorts, the harmonized clinical view was reduced to age at diagnosis and tumour stage, the gene-expression view was restricted to the 476 shared features, and the endpoint was 5-year overall survival rather than 5-year cancer-specific death. The harmonized analysis, therefore, tests whether the copula-based fusion \emph{approach} retains prognostic value out-of-cohort under a shared, deliberately reduced specification, and it should not be read as validating the specific richer clinical model developed in the primary METABRIC analysis. We return to this distinction, and its implications, in the calibration assessment and discussion.

\subsection*{Calibration Assessment}

Discrimination measures such as ROC-AUC quantify how well a score \emph{ranks}
patients but are insensitive to whether predicted probabilities match observed
event frequencies. Because absolute risk interpretation also depends on
\emph{calibration}, we assessed the agreement between predicted and observed
5-year risk for the clinical, gene-expression, and copula-fused scores, both
internally in METABRIC and externally in TCGA. To keep the comparison fair, each score underwent the same second-stage
recalibration procedure. For internal assessment, score-specific logistic
recalibration was estimated by five-fold cross-fitting on the METABRIC
out-of-fold scores, so that each internally evaluated probability was produced
without using that patient to fit the recalibration model. For external
assessment, a final score-specific recalibration model was fitted on all
METABRIC out-of-fold scores and applied unchanged to TCGA. Because the
copula-fused score is a continuous rank-based copula value rather than a
probability, it was mapped to a probability within this recalibration procedure
before any calibration metric was calculated. Thus, no score received a type
of correction that the other scores did not receive.

For each score, we computed the Brier score, the Hosmer--Lemeshow (HL)
goodness-of-fit statistic using ten equal-count bins of predicted risk, the
calibration slope and calibration-in-the-large intercept obtained by logistic
recalibration of the outcome on the score logit, and the integrated calibration
index (ICI), defined as the mean absolute difference between predicted risk and a
non-parametric (isotonic) estimate of observed risk. A calibration slope of one
and intercept of zero indicates ideal calibration; a slope below one indicates
over-dispersed (too extreme) predicted risks, and a non-zero intercept indicates
systematic over- or under-prediction of absolute risk. We complemented these
statistics with reliability curves comparing binned predicted and observed risk.
Because the HL test is known to be highly sensitive to sample size, we interpret
it alongside the ICI, which summarises the \emph{magnitude} of miscalibration on
the probability scale.
%%%%%%%%%%%%%%%%%%%%%%%%%%%%%
As a secondary analysis, we formally compared the external Brier scores using paired patient-level predictions from the TCGA cohort. For two scores \(A\) and \(B\), the paired difference in Brier score was defined as
\[
\Delta_{\mathrm{Brier}}^{A-B}
=
\frac{1}{n}
\sum_{i=1}^{n}
\left[
\left(y_i-\widehat{p}_{i,A}\right)^2
-
\left(y_i-\widehat{p}_{i,B}\right)^2
\right].
\]
A negative value indicates that score \(A\) has a lower Brier score than score \(B\). Ninety-five percent confidence intervals were estimated using \(100{,}000\) paired, outcome-stratified bootstrap resamples of the TCGA evaluation cohort. Two-sided \(p\)-values were obtained using a paired randomization test based on \(500{,}000\) random within-patient exchanges of the two score-specific squared-error losses. To control the familywise error rate, Holm's step-down adjustment was applied jointly to the two comparisons of the copula-fused score with the clinical and gene-expression scores \cite{holm1979simple}. These analyses condition on the fitted METABRIC models and the estimated recalibration functions. Therefore, they quantify sampling uncertainty within the TCGA evaluation cohort but do not propagate uncertainty arising from model development or recalibration estimation.
%%%%%%%%%%%%%%%%%%%%%%%%%%%%%%%%%%%%%%%%%%%%%%%%%%
\subsection*{Feature Importance}
To improve interpretability, we summarised which predictors most influenced each view-specific model. Our primary measure
was repeated five-fold cross-validated permutation importance with ROC-AUC scoring. Within each cross-validation fold, the
fitted pipeline was trained on the training observations, and each predictor was permuted five times in the held-out observations,
yielding $25$ out-of-sample importance estimates per feature. For each feature, we summarised the mean importance, empirical
$2.5$th and $97.5$th percentiles, and the fraction of estimates greater than zero. A feature was classified as stable-positive only
when its empirical $2.5$th percentile was above zero and at least 80\% of its importance estimates were positive. Tree impurity
importance was retained only as a secondary descriptive measure because conventional random-forest importance can be
distorted when predictors are correlated. Agreement between the permutation and impurity rankings was assessed using
Spearman’s $\rho$ and top-$20$ overlap. Biological interpretation was limited to exploratory context for selected higher-ranked genes
and was not treated as evidence of causal or independently stable effects \cite{strobl2007bias,altmann2010permutation}.

%%%%%%%%%%%%%%%%%%%%%%%%
% To improve interpretability, we summarised which predictors most influenced each
% view-specific model. Our primary importance measure was repeated cross-validated
% permutation importance with ROC-AUC scoring: the fitted pipeline was refit within
% each of 5 cross-validation folds, and each predictor was permuted on the held-out
% fold with 5 repeats, giving 25 out-of-sample importance estimates per feature.
% This out-of-sample design is preferred to in-sample permutation, which is
% uninformative for a high-dimensional forest that saturates on its training data.
% For each feature, we summarised the mean importance and flagged features as
% \emph{stable-positive} when the $2.5$th empirical percentile of the estimates was
% above zero and the importance was positive in at least $80\%$ of estimates. Tree
% impurity (Gini) importance was retained only as a secondary descriptive measure,
% and we quantified its agreement with the permutation ranking using Spearman's
% $\rho$ and the top-$20$ overlap, rather than assuming the two rankings coincide.
% Because no gene met the prespecified stable-positive criterion, biological discussion was restricted to a cautious literature-based context
% for selected high-ranked genes and was not presented as evidence of stable individual drivers, confirmed pathways, or causal
% mechanisms.

In the next section, we present the results obtained by applying this approach to the METABRIC and TCGA cohorts.

%%%%%%%%%%%%%%%%%%%%%%%%%%%%%%%%%%%%%%%%%%%%%%%%%%%%%%%%%%%%%

\section*{Results}

\label{sec:results}

\subsection*{Internal evaluation: METABRIC Cohort}
This section presents the main findings from our analysis of the METABRIC cohort. We begin by summarizing the performance of the clinical and gene-expression machine learning risk scores and describing their distributions. We then examine the dependence between the two scores using several copula families, and finally assess whether joint clinical and gene-expression risk strata are associated with differences in long-term outcomes.

\subsection*{Predictive Performance of Clinical and Gene-expression Models}

Here, we summarize the predictive performance of the clinical and gene-expression machine learning models for 5-year cancer-specific death in the METABRIC cohort and describe the distribution of the resulting risk scores. We then move from evaluating each score separately to examining how the two scores relate to one another through copula models.

After applying the 5-year endpoint definition, the analytic cohort included 1,363 patients, of whom 325 experienced cancer-specific death within 5 years, and 1,038 were classified as non-events. Patients who died of other causes or did not have sufficient follow-up for 5-year determination were excluded from this endpoint analysis.

Table~\ref{tab:model_performance} shows the cross-validation ROC-AUC values for each classifier and data view. For the clinical data, all four models performed similarly, with ROC-AUCs ranging from 0.759 to 0.783. The random forest was the best clinical model (AUC 0.783), followed by gradient boosting (AUC 0.766), XGBoost (AUC 0.763), and elastic-net logistic regression (AUC 0.759). For the gene-expression data, performance was lower than for the clinical view but remained clearly better than chance. Again, the random forest performed best (AUC 0.721), followed by XGBoost (AUC 0.695), gradient boosting (AUC 0.677), and elastic-net logistic regression (AUC 0.611). Bootstrap confidence intervals for the best models were 0.783 [0.755, 0.811] for the clinical model and 0.721 [0.690, 0.749] for the gene-expression model. Overall, these results indicate that the clinical predictors provide stronger discrimination for 5-year cancer-specific death, while the gene-expression predictors provide additional, though more modest, prognostic information.
\begin{table}[ht]
\centering
\caption{Cross-validated ROC-AUC for 5-year cancer-specific death prediction by model and view in the METABRIC cohort. Bootstrap 95\% confidence intervals are reported for the best-performing model in each view.}
\label{tab:model_performance}
\begin{tabular}{lccc}
\hline
Model & View & ROC-AUC & 95\% CI \\
\hline
Elastic-net logistic regression & Clinical & 0.759 & --- \\
Random forest & Clinical & \textbf{0.783} & [0.755, 0.811] \\
Gradient boosting & Clinical & 0.766 & --- \\
XGBoost & Clinical & 0.763 & --- \\
Elastic-net logistic regression & Gene-expression & 0.611 & --- \\
Random forest & Gene-expression & \textbf{0.721} & [0.690, 0.749] \\
Gradient boosting & Gene-expression & 0.677 & --- \\
XGBoost & Gene-expression & 0.695 & --- \\
\hline
\end{tabular}
\end{table}

\begin{figure}[ht]
\centering
\includegraphics[width=0.5\textwidth]{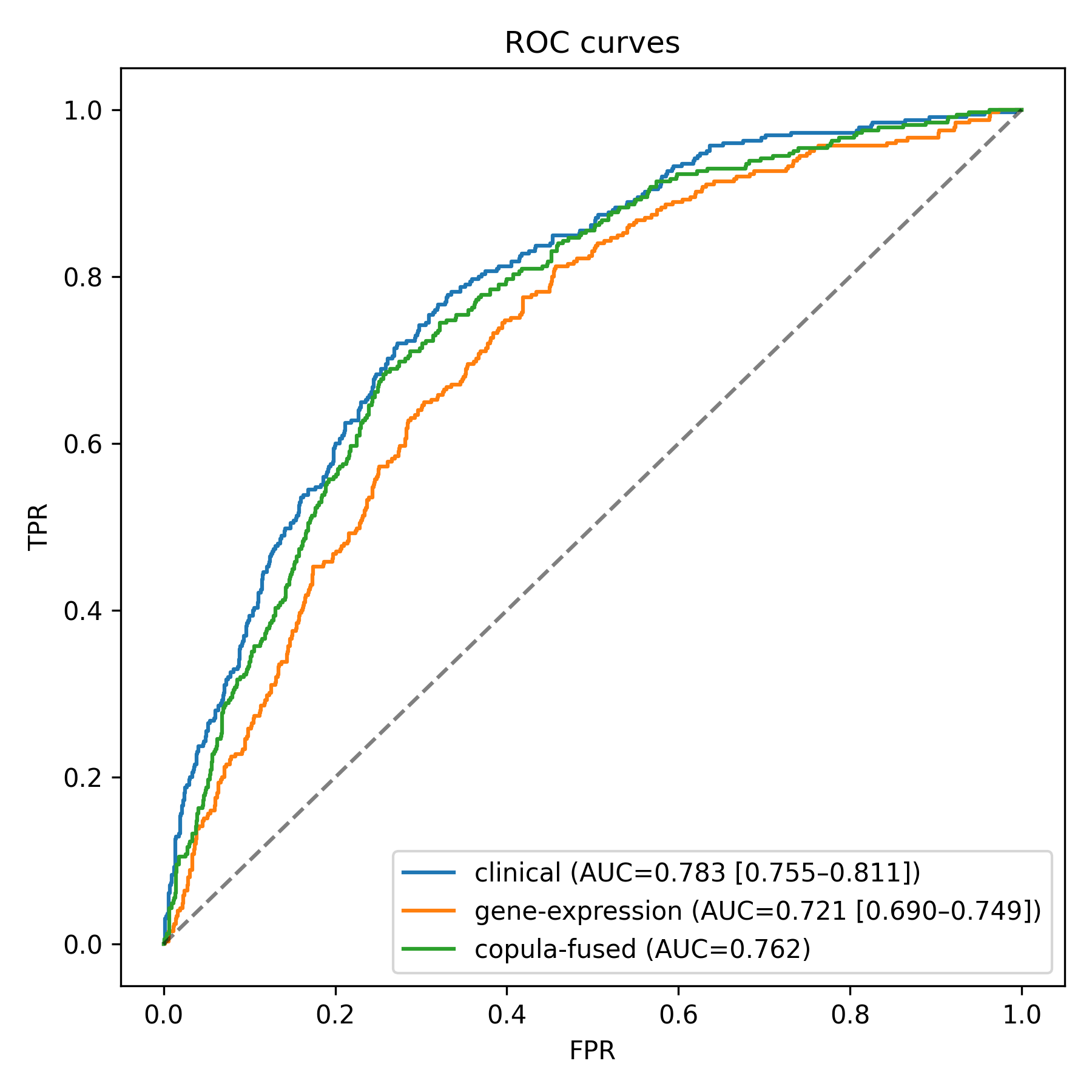}
\caption{ROC curves for 5-year cancer-specific death prediction using the clinical, gene-expression, and copula-fused risk scores in the METABRIC cohort.}
\label{fig:roc}
\end{figure}

Figure~\ref{fig:roc} displays the ROC curves for the best-performing clinical and gene-expression models, together with the copula-fused score. The clinical risk score achieved the highest discrimination (AUC 0.783), the gene-expression score achieved moderate discrimination (AUC 0.721), and the copula-fused score achieved an intermediate AUC of 0.762. Thus, in this cohort, the fused score did not exceed the clinical model in ROC-AUC, but it remained predictive and was useful for downstream joint risk stratification.

\begin{figure}[ht]
\centering
\includegraphics[width=\textwidth]{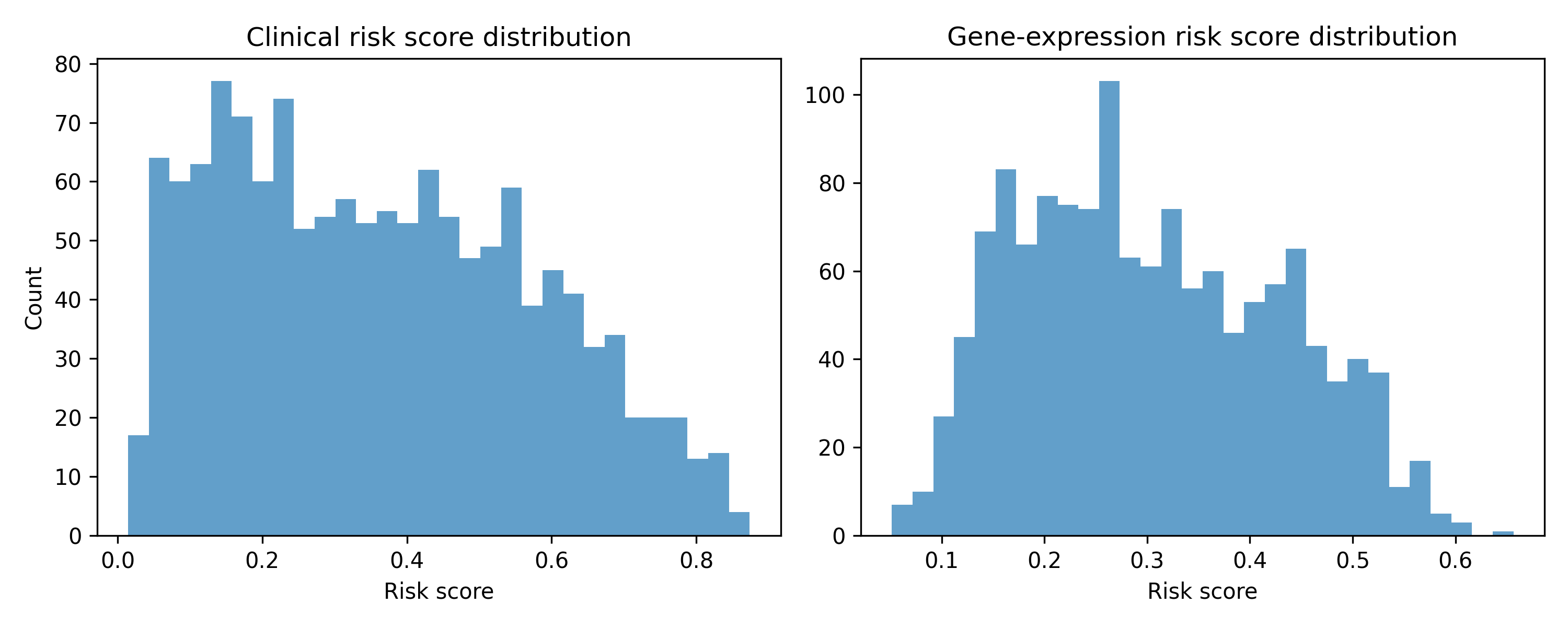}
\caption{Distributions of cross-validated clinical (left) and gene-expression (right) risk scores for 5-year cancer-specific death in the METABRIC cohort.}
\label{fig:risk_hists}
\end{figure}

Figure~\ref{fig:risk_hists} shows the distributions of the cross-validated clinical and gene-expression risk scores. The clinical scores span a broader range, extending from very low values to approximately 0.85, while the gene-expression scores are more concentrated, with most values lying in a narrower middle range. This visual pattern is consistent with the stronger discrimination of the clinical model and the more compressed distribution of the gene-expression score.

To assess whether the gene-expression result depended strongly on the number of molecular features retained, we also performed a sensitivity analysis across several feature counts. The random forest gene-expression model achieved ROC-AUC values of 0.707 with the top 25 features, 0.711 with the top 50, 0.718 with the top 100, 0.718 with the top 200, and 0.721 with all 489 available gene-expression features. These results suggest that the overall gene-expression performance was reasonably stable across a broad range of feature counts, with a slight improvement when more features were retained.

\begin{figure}[ht]
\centering
\includegraphics[width=0.5\textwidth]{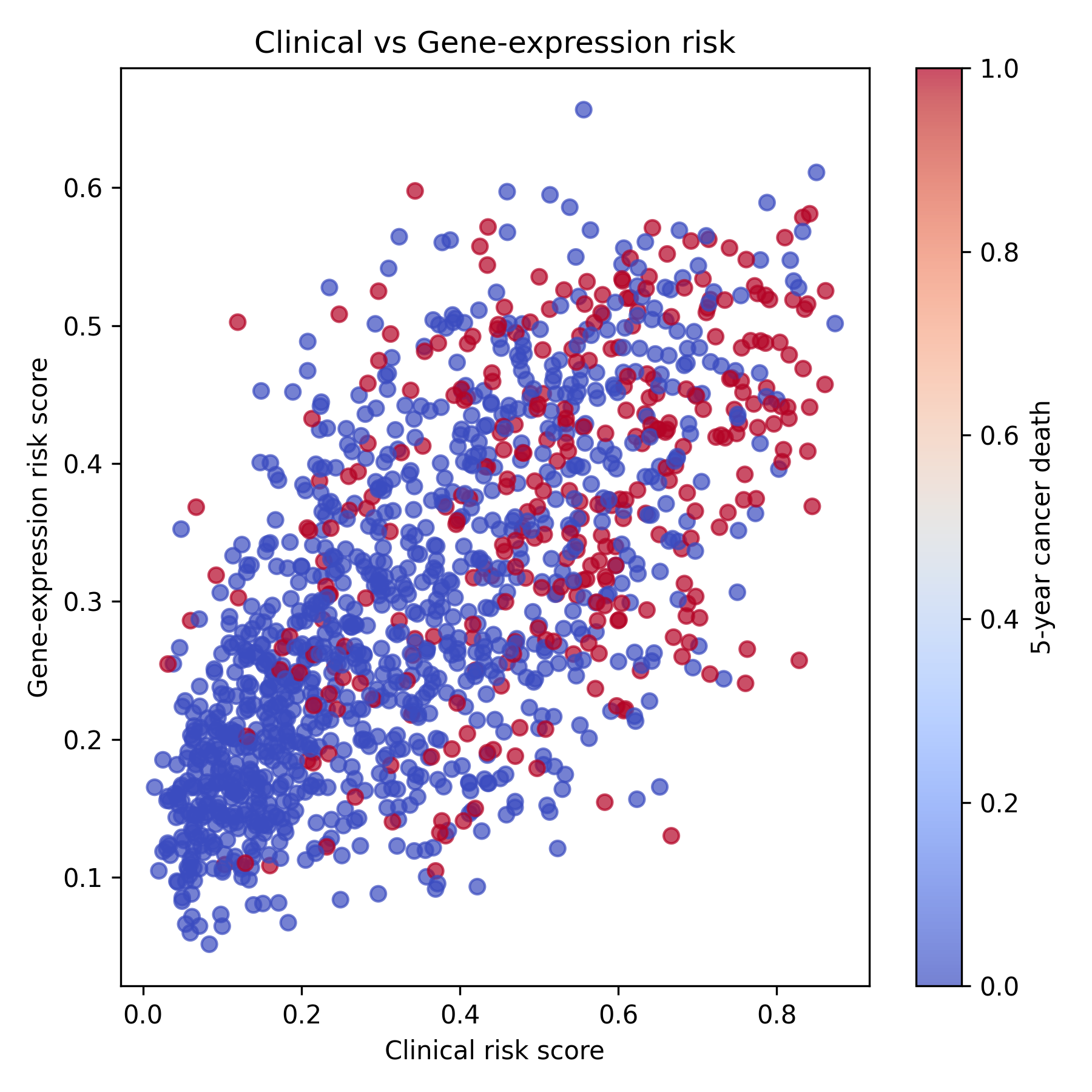}
\caption{Joint scatterplot of cross-validated clinical and gene-expression risk scores, coloured by 5-year cancer-specific death outcome.}
\label{fig:risk_scatter}
\end{figure}

Figure~\ref{fig:risk_scatter} shows a scatterplot of the clinical and gene-expression risk scores, with points coloured by 5-year outcome. The two scores are positively associated, forming an upward-sloping cloud. Patients with 5-year cancer death are more commonly found toward the upper-right region, where both scores are elevated, while non-events are more concentrated toward the lower-left. This pattern motivates the copula analysis below, in which we model the joint distribution of the two scores directly rather than treating each score in isolation.

Since both the clinical and gene-expression models provide prognostic signal, and their risk scores are positively related, we next examine their dependence structure more formally using copula models.

\subsection*{Copula-based Dependence Between Clinical and Gene-expression Risk Scores}

In this part, we examine how the clinical and gene-expression risk scores vary together across their range. We used pseudo-observations derived from the ranked scores to fit four parametric copula families: Gaussian, Clayton, Gumbel, and Frank.

Table~\ref{tab:tail_dep} summarizes the fitted parameters and tail-dependence coefficients. The Gaussian copula, with correlation parameter $\rho = 0.682$, implies symmetric dependence without asymptotic tail dependence. The Clayton copula, with $\theta = 1.832$, captures lower-tail dependence ($\lambda_L = 0.685$, $\lambda_U = 0$), whereas the Gumbel copula, with $\theta = 1.916$, captures upper-tail dependence ($\lambda_U = 0.564$, $\lambda_L = 0$). The Frank copula, with parameter 5.352, represents positive dependence without asymptotic tail dependence in either tail.

\begin{table}[ht]
\centering
\caption{Fitted copula parameters, bootstrap 95\% confidence intervals, and tail-dependence coefficients for the joint distribution of clinical and gene-expression risk scores.}
\label{tab:tail_dep}
\begin{tabular}{lcccc}
\hline
Copula & Parameter & 95\% CI & $\lambda_L$ & $\lambda_U$ \\
\hline
Gaussian & $\rho = 0.682$ & [0.652, 0.711] & 0.000 & 0.000 \\
Clayton & $\theta = 1.832$ & [1.652, 2.026] & 0.685 & 0.000 \\
Gumbel & $\theta = 1.916$ & [1.826, 2.013] & 0.000 & 0.564 \\
Frank & $\theta = 5.352$ & [4.931, 5.795] & 0.000 & 0.000 \\
\hline
\end{tabular}
\end{table}

To compare goodness-of-fit, we used the Cramér--von Mises statistic with a parametric bootstrap. Table~\ref{tab:gof} shows the results. The Frank copula gave the smallest Cramér--von Mises statistic, $3.1\times10^{-5},$ and was therefore selected as the best-fitting family according to the minimum-statistic criterion, followed very closely by the Gaussian copula, $3.3\times10^{-5}.$ The corresponding bootstrap $p$-values, $0.955$ for the Frank copula and $0.952$ for the Gaussian copula, confirmed that both families were adequate for the observed dependence. In contrast, the Clayton and Gumbel copulas showed larger discrepancies. These results suggest that the dependence between the clinical and gene-expression risk scores is positive and moderate. Although the Frank copula achieved the smallest goodness-of-fit statistic, its fit was only marginally better than that of the Gaussian copula. Therefore, the results do not provide strong evidence that the Frank copula fits substantially better than all competing families. Nevertheless, the Frank copula was retained for the primary analysis because it produced the smallest Cramér--von Mises statistic among the copula families considered.

% To compare goodness-of-fit, we used the Cramér--von Mises statistic with a parametric bootstrap. Table~\ref{tab:gof} shows the results. The Frank copula gave the smallest Cram\'er--von Mises statistic ($3.1\times10^{-5}$) and was therefore selected as the best-fitting family, followed very closely by the Gaussian copula ($3.3\times10^{-5}$). The corresponding bootstrap $p$-values (Frank $0.955$, Gaussian $0.952$) confirmed that both families were adequate for the observed dependence. In contrast, the Clayton and Gumbel copulas showed larger discrepancies. These results suggest that the dependence between the clinical and gene-expression risk scores is positive and moderate, and is better described by a Frank copula than by the other families considered here.

\begin{table}[ht]
\centering
\caption{Cramér--von Mises goodness-of-fit statistics and bootstrap $p$-values for fitted copulas in the METABRIC cohort.}
\label{tab:gof}
\begin{tabular}{lcc}
\hline
Copula & Cramér--von Mises statistic & Bootstrap $p$-value \\
\hline
Gaussian & $3.3 \times 10^{-5}$ & 0.952 \\
Clayton & $1.87 \times 10^{-4}$ & 0.153 \\
Gumbel & $1.23 \times 10^{-4}$ & 0.312 \\
Frank & $3.1 \times 10^{-5}$ & 0.955 \\
\hline
\end{tabular}
\end{table}

The numerical results are consistent with the graphical comparisons. The empirical copula heat map and the fitted Frank copula heat map are visually very similar, showing a smooth increase in joint probability from the lower-left to the upper-right of the unit square. Likewise, the contour plots show that the fitted Frank copula follows the overall diagonal dependence pattern of the pseudo-observations. For completeness, we also examined Gaussian, Clayton, and Gumbel heat maps and contour overlays, which showed broadly similar overall structure but somewhat weaker goodness-of-fit than the Frank model.

\begin{figure}[ht]
\centering
\includegraphics[width=\textwidth]{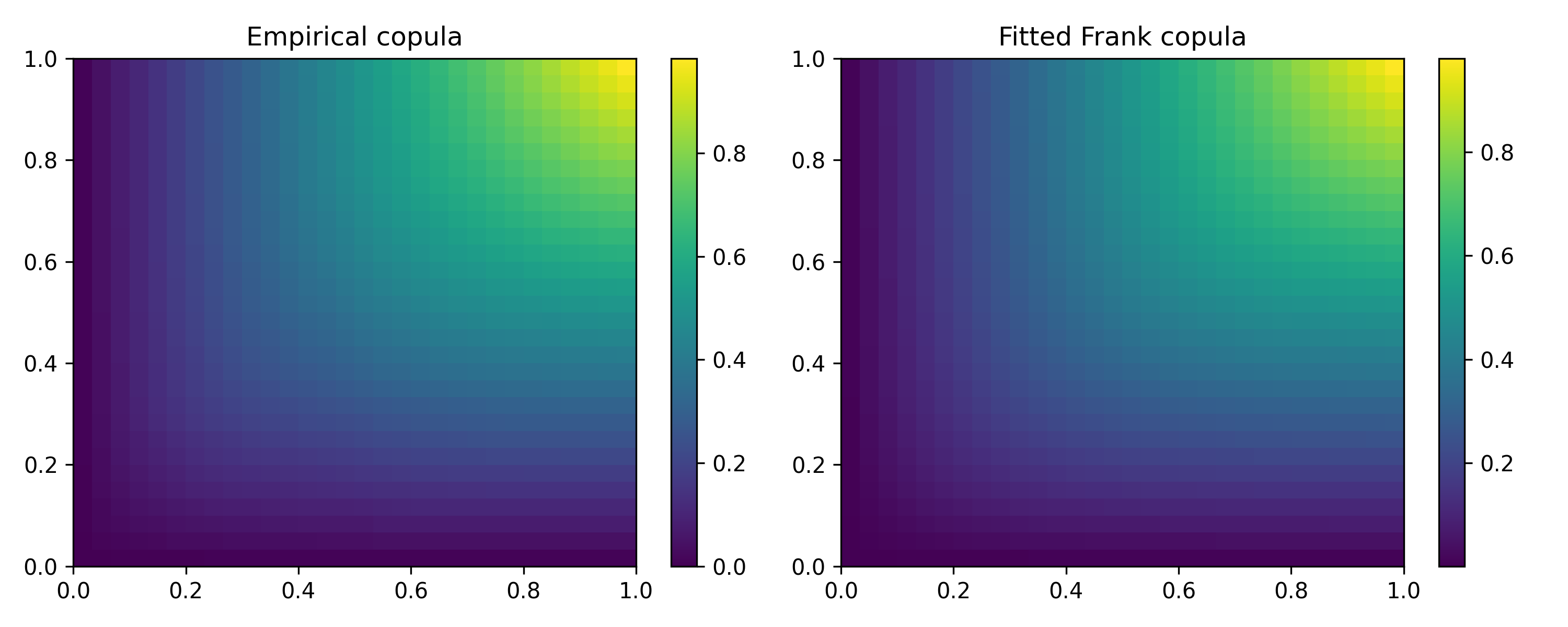}
\caption{Empirical copula of the pseudo-observations (left) and fitted Frank copula (right) for the joint distribution of clinical and gene-expression risk scores.}
\label{fig:copula_heat}
\end{figure}

\begin{figure}[ht]
\centering
\includegraphics[width=0.5\textwidth]{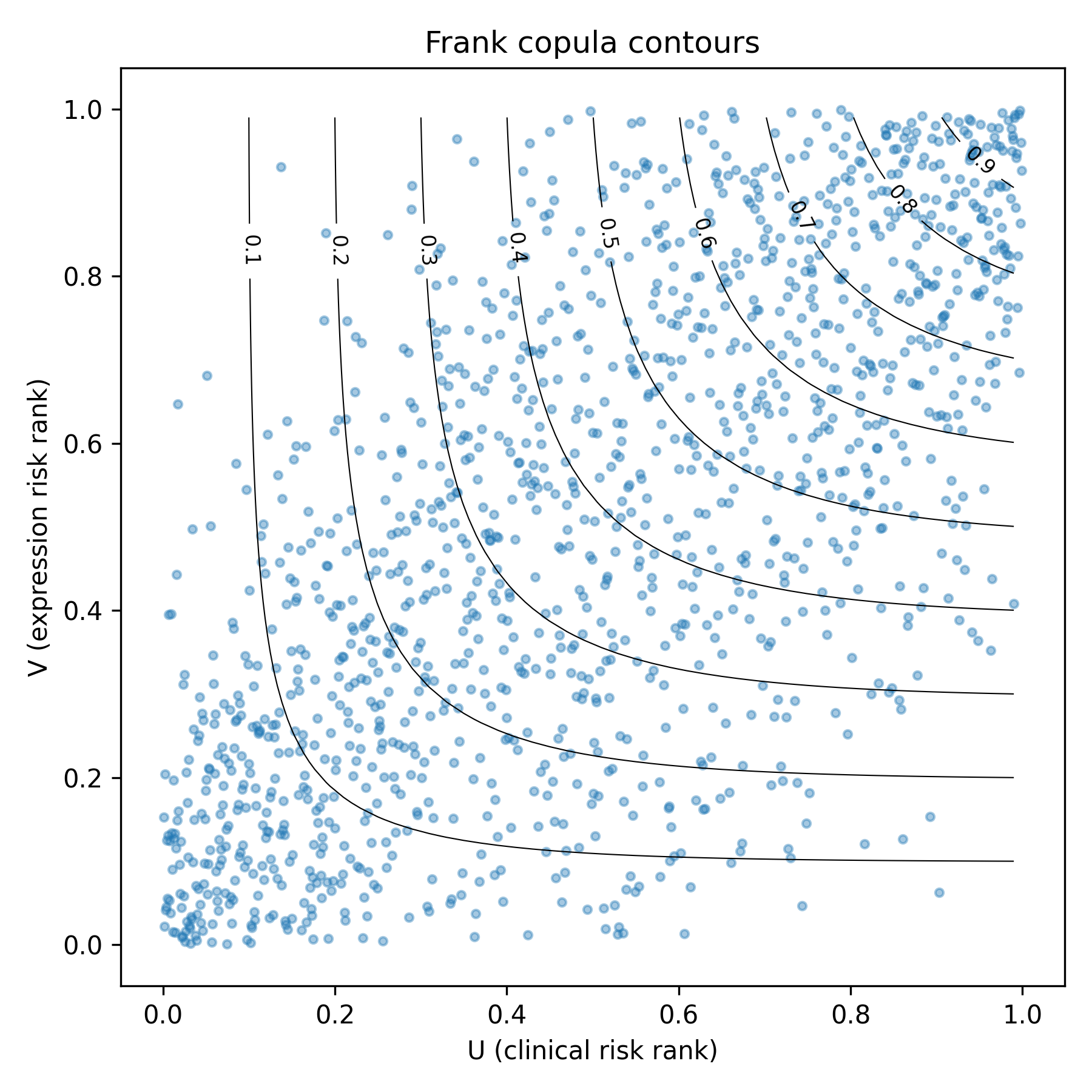}
\caption{Frank copula contours overlaid on the pseudo-observations derived from the clinical and gene-expression risk scores.}
\label{fig:copula_contours}
\end{figure}

Using the best-fitting Frank copula, we then formed a copula-based joint risk score. This fused score achieved a ROC-AUC of 0.762 with a bootstrap 95\% confidence interval of [0.734, 0.790]. Although this did not exceed the clinical model’s ROC-AUC, it remained predictive and supported the joint risk-stratification analyses described next.

To test whether the copula contributes to discrimination beyond a trivial score
combination, we compared the copula-fused score against three simple
score-level fusion baselines: the rank average of the two pseudo-observations,
the average of the two predicted probabilities, and a logistic-regression stacker
of the two out-of-fold scores (itself evaluated by 5-fold cross-validation).
As shown in Table~\ref{tab:fusion_baselines}, these baselines matched or slightly
exceeded the copula-fused score in the internal cohort (rank-average $0.776$,
probability-average $0.784$, logistic-stacking $0.783$, versus copula-fused
$0.762$), and all fusion scores had confidence intervals overlapping that of the
clinical model. The copula-fused score therefore, does not provide a discrimination
advantage over simple fusion; consistent with the aims of this study, the value of
the copula lies in the interpretable characterization of the dependence structure
(Tables~\ref{tab:tail_dep}--\ref{tab:gof}) and in the joint risk stratification
described below, rather than in improved ranking performance.

\begin{table}[ht]
\centering
\caption{Discrimination (ROC-AUC, 95\% bootstrap CI) of the copula-fused score
compared with the individual clinical and gene-expression scores and with three
simple score-level fusion baselines, internally in METABRIC (5-year
cancer-specific death) and externally in TCGA (5-year overall survival).}
\label{tab:fusion_baselines}
\begin{tabular}{lcc}
\hline
Score & METABRIC, AUC (95\% CI) & TCGA, AUC (95\% CI) \\
\hline
Clinical               & 0.783 [0.755, 0.811] & 0.671 [0.598, 0.735] \\
Gene-expression        & 0.721 [0.690, 0.749] & 0.639 [0.572, 0.704] \\
Copula-fused           & 0.762 [0.734, 0.790] & 0.696 [0.626, 0.759] \\
Rank-average           & 0.776 [0.749, 0.802] & 0.709 [0.641, 0.770] \\
Probability-average    & 0.784 [0.758, 0.810] & 0.702 [0.631, 0.761] \\
Logistic-stacking      & 0.783 [0.756, 0.810] & 0.706 [0.638, 0.767] \\
\hline
\end{tabular}
\end{table}

Although the best-fitting Frank and Gaussian copulas imply symmetric dependence
without asymptotic tail dependence, the Gumbel copula yielded a non-negligible
upper-tail dependence coefficient ($\lambda_U = 0.564$; Table~\ref{tab:tail_dep}).
We interpret this as an exploratory observation rather than a primary finding,
because the Gumbel copula did not provide the best fit to the joint distribution
(Table~\ref{tab:gof}). Nonetheless, 
the Gumbel fit suggested a possible upper-tail pattern, although it did not have the smallest discrepancy. This exploratory observation
does not support treatment intensification or clinical risk classification and should be evaluated in larger cohorts using models designed to assess upper-tail dependence.
% the possibility of upper-tail co-occurrence,
% in which patients at extreme clinical risk also tend to be at extreme
% gene-expression risk is clinically relevant, since such concordant
% extreme-risk patients may be of particular interest for intensified management.
We therefore indicate this as a hypothesis warranting dedicated investigation in
larger cohorts and with copula families specifically designed to capture
upper-tail behaviour, rather than as a conclusion supported by the present
best-fitting model.

\subsection*{Joint Risk Strata and Survival}

Here, we examine whether combining the clinical and gene-expression risk scores identifies groups with different long-term outcomes. We used the sample medians of the two scores to define four joint risk strata: low--low, high-clinical-only, high-expression-only, and high-both. The strata cutoffs were the medians of the analytic out-of-fold scores (clinical $0.344$, gene-expression $0.287$). For the survival analyses, we applied these strata to the full METABRIC cohort ($n=1{,}904$): the $1{,}363$ analytic patients retained their out-of-fold scores, while the $541$ patients excluded from the binary endpoint (and therefore never used to train any model) received out-of-sample predictions from the final fitted models. On the full cohort this yielded $740$ low--low, $280$ high-clinical-only, $232$ high-expression-only, and $652$ high-both patients.

\begin{figure}[ht]
\centering
\includegraphics[width=0.5\textwidth]{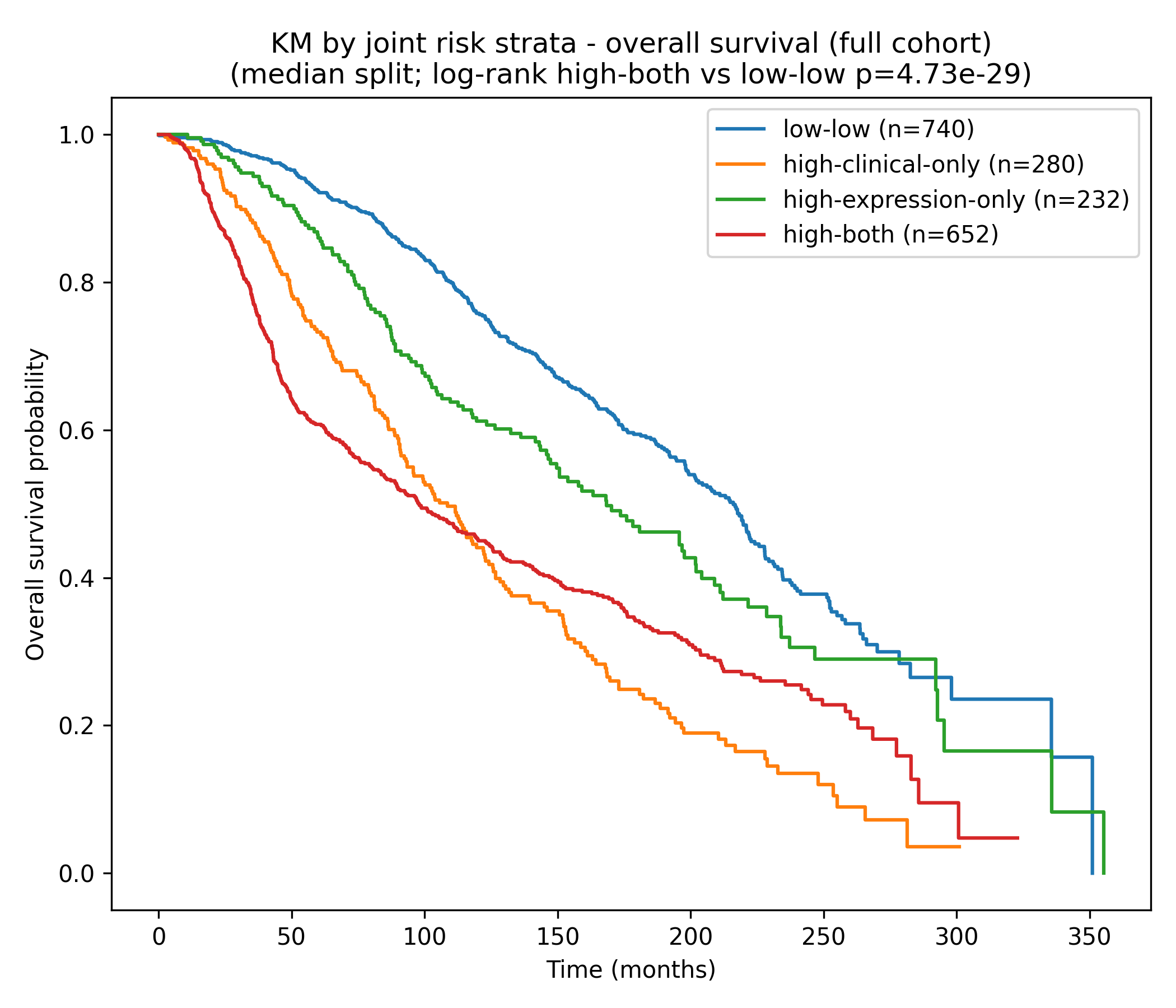}
\caption{Kaplan--Meier curves for overall (all-cause) survival by joint risk group on the full METABRIC cohort ($n=1{,}904$), with risk strata defined using the medians of the analytic out-of-fold clinical and gene-expression risk scores.}
\label{fig:km_joint}
\end{figure}

Figure~\ref{fig:km_joint} presents the Kaplan--Meier curves for overall (all-cause) survival on the full cohort across these four groups; a death from any cause was treated as an event. The low--low group shows the most favorable survival over follow-up. Patients in the high-clinical-only and high-expression-only groups have poorer survival than the low--low group, indicating that elevation in either score alone is associated with worse outcomes. The high-both group shows the least favorable survival overall, with its curve dropping earlier and remaining below the low--low curve throughout follow-up. The log-rank comparison between the high-both and low--low groups was highly significant ($p = 4.7 \times 10^{-29}$).

\begin{figure}[ht]
\centering
\includegraphics[width=0.6\textwidth]{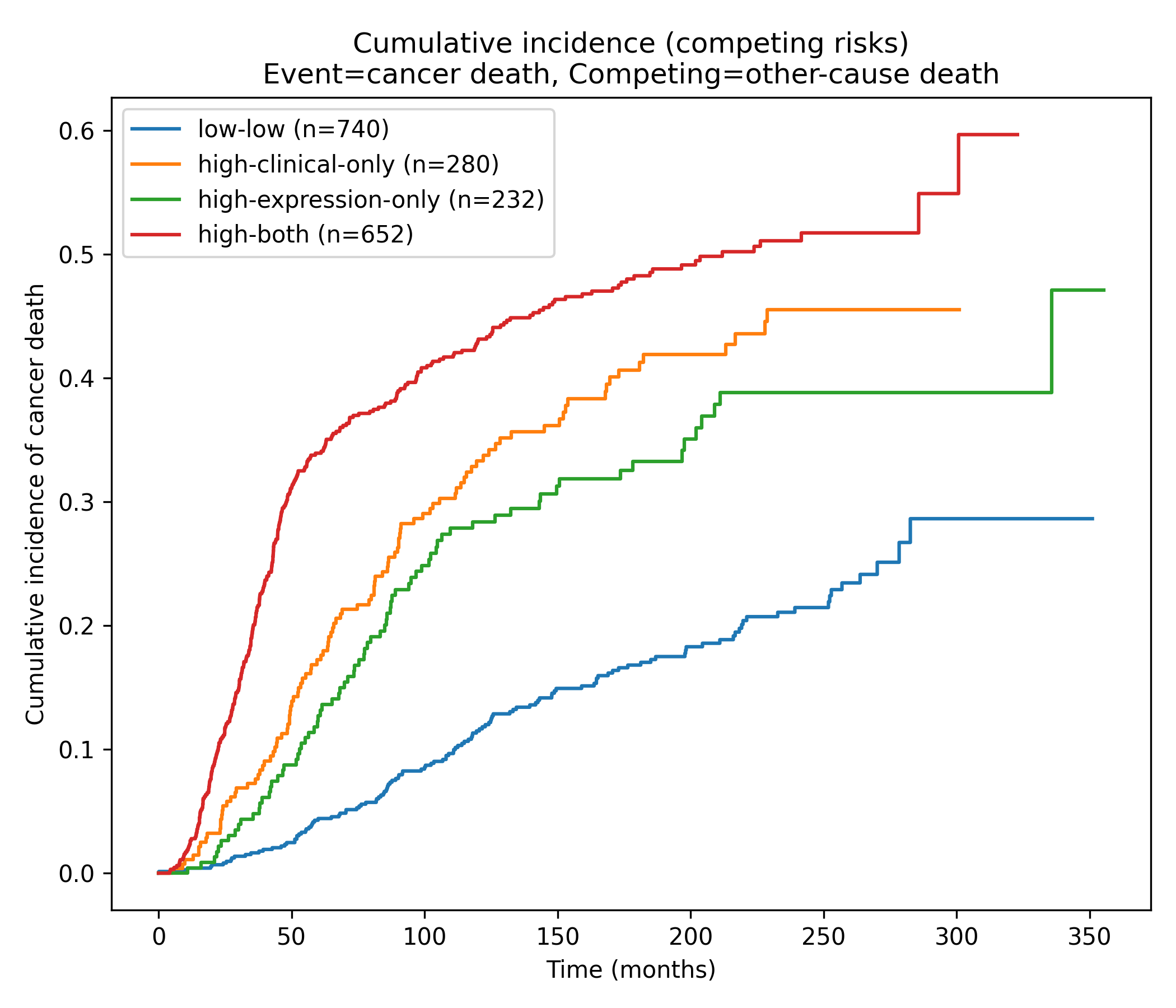}
\caption{Cumulative incidence of cancer death by joint risk group under a competing-risks framework, with death from other causes treated as the competing event. The risk-group cutoffs and group assignments are identical to those used in the Kaplan--Meier analysis (Figure~\ref{fig:km_joint}).}
\label{fig:cif}
\end{figure}

Because deaths from causes other than breast cancer were explicitly excluded from the 5-year cancer-specific endpoint, we also examined cumulative incidence under a competing-risks framework using the full cohort. To avoid in-sample optimism in the definition of the risk groups, patients in the analytic cohort were stratified using their out-of-fold risk scores, and the remaining patients (who were excluded from the endpoint and therefore never used to train any model) were scored with genuine out-of-sample predictions; the resulting risk strata are identical to those used in the Kaplan--Meier analysis above ($740$/$280$/$232$/$652$). In that analysis, there were 622 cancer deaths, 480 other-cause deaths, and 802 censored observations. Figure~\ref{fig:cif} shows that the cumulative incidence of cancer death was highest in the high-both group, intermediate in the high-clinical-only group, and lower in the low--low and high-expression-only groups over much of follow-up. The cause-specific log-rank comparison between the high-both and low--low groups was highly significant ($p = 6.6 \times 10^{-43}$).

Overall, these results show that the clinical model provides the strongest discrimination for 5-year cancer-specific death in METABRIC, while the gene-expression model contributes additional prognostic structure. The two scores are positively dependent, and among the copula families considered here, the Frank copula provided the best fit to their joint distribution. Although the copula-fused score did not improve upon the clinical model in ROC-AUC, the joint clinical and gene-expression stratification still identified groups with clearly different long-term survival and cumulative incidence patterns.

\subsection*{External Evaluation: TCGA Cohort}

To assess out-of-cohort generalizability, we evaluated the harmonized clinical, gene-expression, and copula-fused risk scores in an independent TCGA cohort after model development in METABRIC. The harmonization procedure retained the shared clinical predictors and 476 shared gene-expression features across the two cohorts. After applying the common 5-year overall survival endpoint, the harmonized METABRIC dataset comprised 1,844 patients with 412 events, whereas the TCGA evaluation cohort comprised 317 patients with 85 events.

Within the harmonized METABRIC training cohort, the best-performing clinical model was XGBoost, while the best-performing gene-expression model was random forest. The internal performance estimates from the harmonized external-evaluation pipeline are not directly comparable with those from the primary METABRIC analysis because the harmonized pipeline used a reduced predictor set and a different endpoint. Specifically, the clinical model was limited to age at diagnosis and tumor stage, the gene-expression model was limited to the 476 features shared between METABRIC and TCGA, and the outcome was 5-year overall survival rather than 5-year cancer-specific death. These design differences may contribute to the lower internal clinical AUC observed in the harmonized pipeline (0.664 versus 0.783). The internal ROC-AUC for the clinical score was 0.664 with a 95\% bootstrap confidence interval of [0.635, 0.692]. The internal gene-expression score achieved a ROC-AUC of 0.689 [0.663, 0.718]. After copula-based fusion, the internal fused score achieved the highest discrimination, with ROC-AUC 0.724 [0.699, 0.751].

For the dependence modelling step in the harmonized training pipeline, the Gaussian copula had the smallest Cram\'er--von Mises statistic and was selected on that basis. The fit was, however, a near-tie with the Clayton copula (Cram\'er--von Mises $9.66\times10^{-6}$ versus $9.71\times10^{-6}$, about $0.5\%$ apart), so the family is not uniquely determined by the data. The fitted Gaussian parameter was $\rho = 0.141$, indicating a weak positive dependence between the harmonized clinical and gene-expression training scores.

We then applied the final trained METABRIC models unchanged to the independent TCGA cohort. In external evaluation, the clinical score achieved a ROC-AUC of 0.671 with 95\% bootstrap confidence interval [0.598, 0.735], while the gene-expression score achieved a ROC-AUC of 0.639 [0.572, 0.704]. Among these three scores, the copula-fused score achieved the highest external ROC-AUC, 0.696 [0.626, 0.759], showing slightly better discrimination than either individual score. However, as in the internal cohort, the simple fusion baselines performed comparably or marginally better than the copula in TCGA (rank-average $0.709$, logistic-stacking $0.706$, probability-average $0.702$; Table~\ref{tab:fusion_baselines}), and all six scores had substantially overlapping confidence intervals spanning roughly $0.63$--$0.77$. We therefore do not interpret the external results as evidence that copula fusion improves discrimination relative to trivial score combination; rather, they indicate that the fused score retained a level of out-of-cohort discrimination comparable to both the individual scores and to simple fusion, while additionally providing an interpretable dependence model and, after recalibration, favorable calibration (below).

\begin{figure}[ht]
\centering
\includegraphics[width=0.5\textwidth]{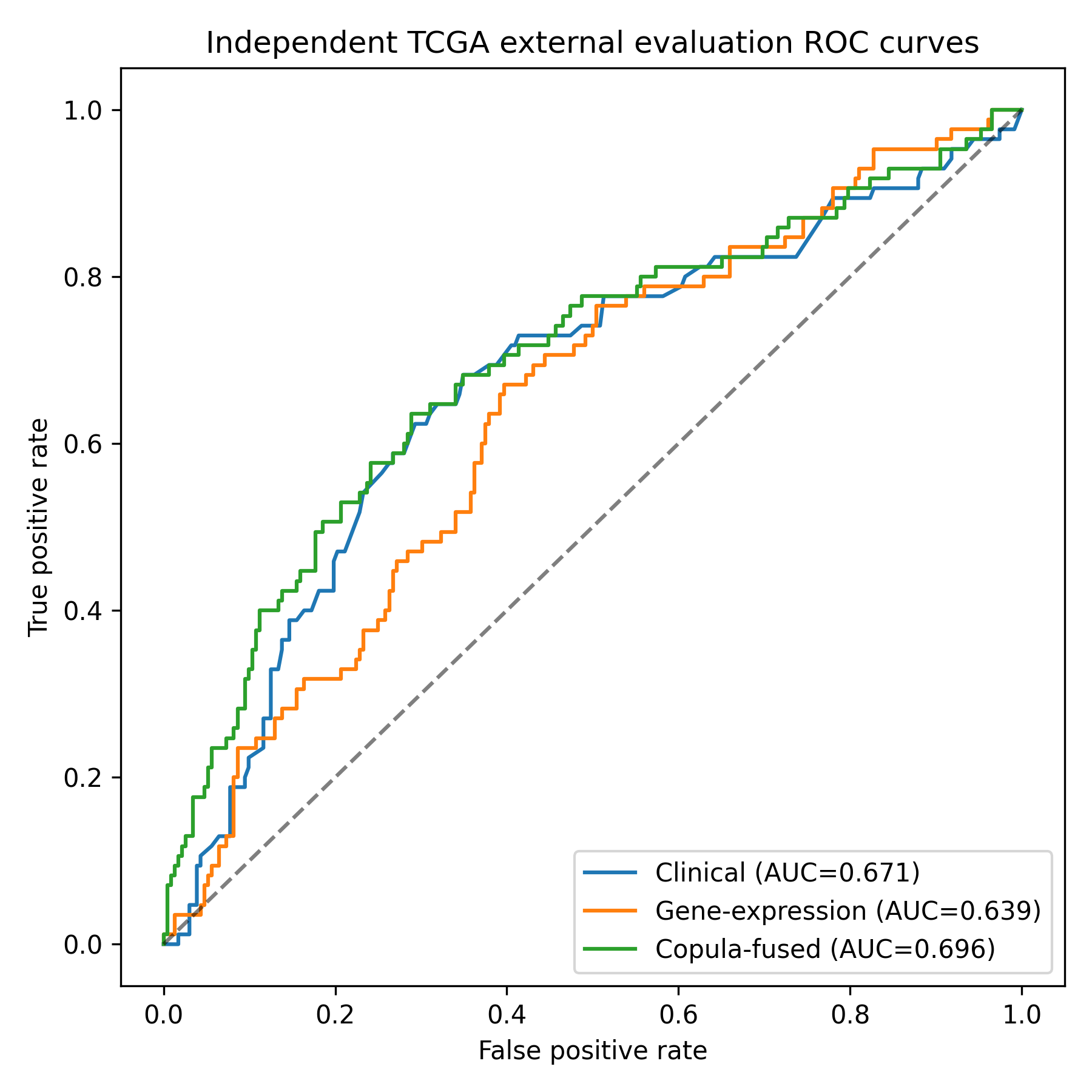}
\caption{External evaluation ROC curves in the TCGA cohort for the harmonized clinical, gene-expression, and copula-fused risk scores.}
\label{fig:external_roc}
\end{figure}

Table~\ref{tab:external_evaluation_auc} summarizes the internal METABRIC and external TCGA ROC-AUC values and corresponding bootstrap confidence intervals for the three scores. Figure~\ref{fig:external_roc} shows the external ROC curves in TCGA. The fused score remains above the reference line and is modestly higher than the two individual scores across much of the ROC space, consistent with a small but favorable gain in external discrimination.

\begin{table}[ht]
\centering
\caption{Internal METABRIC and external TCGA ROC-AUC values with 95\% bootstrap confidence intervals for the harmonized clinical, gene-expression, and copula-fused risk scores.}

\label{tab:external_evaluation_auc}
\begin{tabular}{lcc}
\hline
Score & METABRIC, AUC (95\% CI) & TCGA, AUC (95\% CI) \\
\hline
Clinical & 0.664 [0.635, 0.692] & 0.671 [0.598, 0.735] \\
Gene-expression & 0.689 [0.663, 0.718] & 0.639 [0.572, 0.704] \\
Copula-fused & 0.724 [0.699, 0.751] & 0.696 [0.626, 0.759] \\
\hline
\end{tabular}
\end{table}

\subsection*{Calibration Assessment}

Because ROC-AUC reflects only ranking, we separately assessed how well the
predicted risks matched observed 5-year event frequencies. To keep the
comparison fair, each score underwent the same second-stage recalibration
procedure. Internal METABRIC calibration was evaluated using score-specific,
five-fold cross-fitted recalibrated probabilities. For external evaluation,
final score-specific recalibration models were fitted on all METABRIC
out-of-fold scores and applied unchanged to TCGA. The copula-fused score, which
is a continuous rank-based copula value rather than a probability, was mapped
to a probability within this procedure. Table~\ref{tab:calibration} reports the
Brier score, Hosmer--Lemeshow (HL) statistic, calibration slope and intercept,
and integrated calibration index (ICI) for all three recalibrated scores under
the 5-year overall-survival specification.

% After this identical recalibration, internal calibration in METABRIC was
% adequate for the clinical and gene-expression scores (HL $p=0.17$ and $0.16$;
% slopes $0.94$ and $0.98$; ICI $0.022$ and $0.018$) and weaker for the
% copula-fused score (HL $p=0.003$; ICI $0.039$), although the absolute
% miscalibration of all three scores was small. In the independent TCGA cohort, the recalibrated copula-fused score showed the most favorable calibration point estimates among the three scores, with an HL $p$-value of $0.79$, a calibration slope of $0.99$, an ICI of $0.035$, and the lowest Brier score of $0.174$. The clinical and gene-expression scores had HL $p$-values of $0.079$ and $0.050$, respectively. This external pattern contrasts with the internal METABRIC results, where the copula-fused score was not the best calibrated. These findings suggest favorable external calibration of the fused score after identical recalibration of all three scores. However, because uncertainty intervals and formal pairwise comparisons were not calculated for these calibration measures, we do not interpret the results as establishing a definitive calibration advantage. The HL statistic was interpreted alongside the ICI because the HL test is sensitive to sample size.
After this identical recalibration, internal calibration in METABRIC was adequate for the clinical and gene-expression scores (HL \(p=0.17\) and \(0.16\), slopes \(0.94\) and \(0.98\), and ICI values \(0.022\) and \(0.018\), respectively) and weaker for the copula-fused score (HL \(p=0.003\) and ICI \(0.039\)), although the absolute miscalibration estimates were small. In the independent TCGA cohort, the recalibrated copula-fused score showed the most favourable calibration point estimates among the three scores, with an HL \(p\)-value of \(0.79\), a calibration slope of \(0.99\), an ICI of \(0.035\), and the lowest Brier score of \(0.174\). The corresponding Brier scores were \(0.186\) for the clinical score and \(0.190\) for the gene-expression score.

Using the unrounded patient-level probabilities, the paired Brier-score difference between the copula-fused and clinical scores was \(\Delta_{\mathrm{Brier}}^{F-C}=-0.0118\) (\(95\%\,\mathrm{CI}: -0.0240,\ 0.0007\); \(p_{\mathrm{Holm}}=0.065\)). Thus, the copula-fused score had a numerically lower Brier score than the clinical score, but the difference did not reach the conventional \(0.05\) significance level. The paired difference between the copula-fused and gene-expression scores was \(\Delta_{\mathrm{Brier}}^{F-G}=-0.0154\) (\(95\%\,\mathrm{CI}: -0.0254,\ -0.0050\); \(p_{\mathrm{Holm}}=0.021\)), providing evidence of a lower Brier score for the copula-fused score than for the gene-expression score. The calibration slope and ICI point estimates also favoured the copula-fused score, but formal pairwise inference was not performed for these measures. The HL \(p\)-values were interpreted as model-specific goodness-of-fit diagnostics rather than as comparative tests of calibration.

% In the independent TCGA cohort the
% pattern favoured the fused score: the recalibrated copula-fused score was the 
% best calibrated of the three (HL $p=0.79$; calibration slope $0.99$; ICI $0.035$;
% lowest Brier score, $0.174$), while the clinical and gene-expression scores were
% adequate to marginal (HL $p=0.079$ and $0.050$). Thus, when every score is
% recalibrated identically, the fused score's calibration transports best to the
% external cohort, even though internally it is not the best calibrated. We
% interpret the HL statistic alongside the ICI because the HL test is highly
% sensitive to sample size.

For completeness, we also assessed calibration in the primary cancer-specific
analysis. The raw discrimination-optimized probabilities over-predicted absolute
risk (clinical calibration intercept $-0.74$, ICI $0.127$; gene-expression
intercept $-0.36$, ICI $0.066$), consistent with the effect of balanced class
weighting on a cohort with a $24\%$ event rate; after the same cross-fitted
recalibration, all three scores were well calibrated (clinical HL $p=0.74$,
ICI $0.026$; gene-expression HL $p=0.25$, ICI $0.027$; copula-fused HL $p=0.45$,
ICI $0.028$). Taken together, cross-fitted recalibration improved absolute-risk calibration in these analyses, while the specific contributions of the copula framework were the explicit description of score dependence and the descriptive joint score-group analysis.

% Taken together, the calibration results indicate that the
% framework's value lies primarily in discrimination and risk
% \emph{stratification}, and that cross-fitted recalibration substantially improved
% calibration in these analyses, although additional prospective calibration
% assessment would be required before individual-level risk communication; this is
% consistent with our framing of the study as methodological and exploratory rather
% than deployment-ready.

\begin{table}[ht]
\centering
\caption{Calibration of the clinical, gene-expression, and copula-fused scores
under the 5-year overall-survival specification, internally in METABRIC and
externally in TCGA. All three scores underwent the same second-stage logistic
recalibration procedure. Internal estimates use five-fold cross-fitted
recalibrated probabilities from METABRIC. External estimates use score-specific
recalibration models fitted on all METABRIC out-of-fold scores and applied
unchanged to TCGA. HL: Hosmer--Lemeshow; ICI: integrated calibration index.
Ideal calibration corresponds to slope $=1$, intercept $=0$, and small Brier
and ICI values. Pairwise Brier-score differences were calculated from the
unrounded patient-level TCGA probabilities and are reported in the text.
Negative paired differences indicate a lower Brier score for the copula-fused
score.}
\label{tab:calibration}
\begin{tabular}{llccccc}
\hline
Cohort & Score & Brier & HL $p$-value & Slope & Intercept & ICI \\
\hline
METABRIC (internal) & Clinical        & 0.162 & 0.171 & 0.943 & $-0.002$ & 0.022 \\
METABRIC (internal) & Gene-expression & 0.161 & 0.162 & 0.979 & $-0.000$ & 0.018 \\
METABRIC (internal) & Copula-fused    & 0.157 & 0.003 & 0.978 & $0.001$  & 0.039 \\
TCGA (external)     & Clinical        & 0.186 & 0.079 & 0.710 & $-0.059$ & 0.052 \\
TCGA (external)     & Gene-expression & 0.190 & 0.050 & 0.913 & $0.282$  & 0.053 \\
TCGA (external)     & Copula-fused    & 0.174 & 0.792 & 0.991 & $0.057$  & 0.035 \\
\hline
\end{tabular}
\end{table}

\begin{figure}[ht]
\centering
\includegraphics[width=0.49\textwidth]{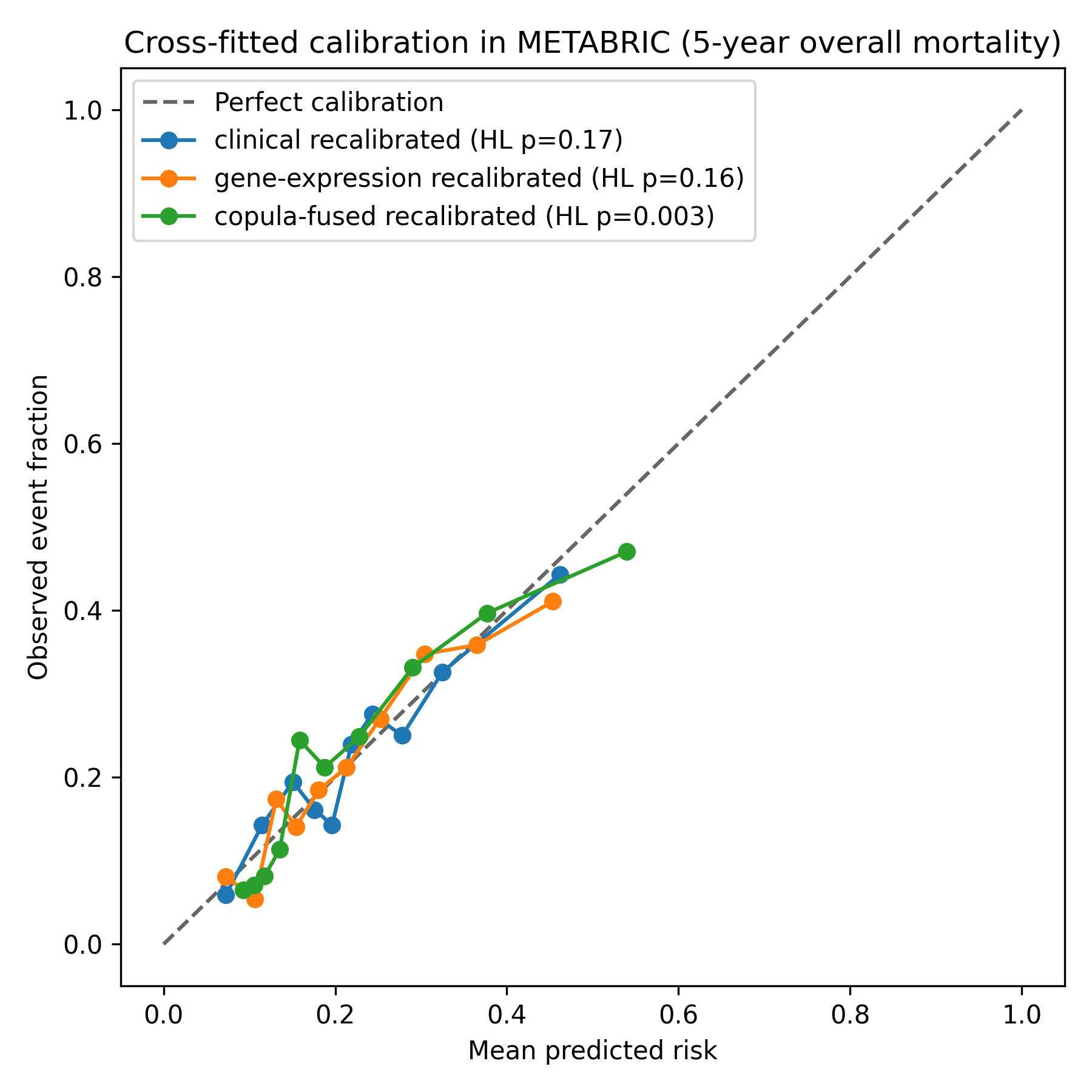}
\includegraphics[width=0.49\textwidth]{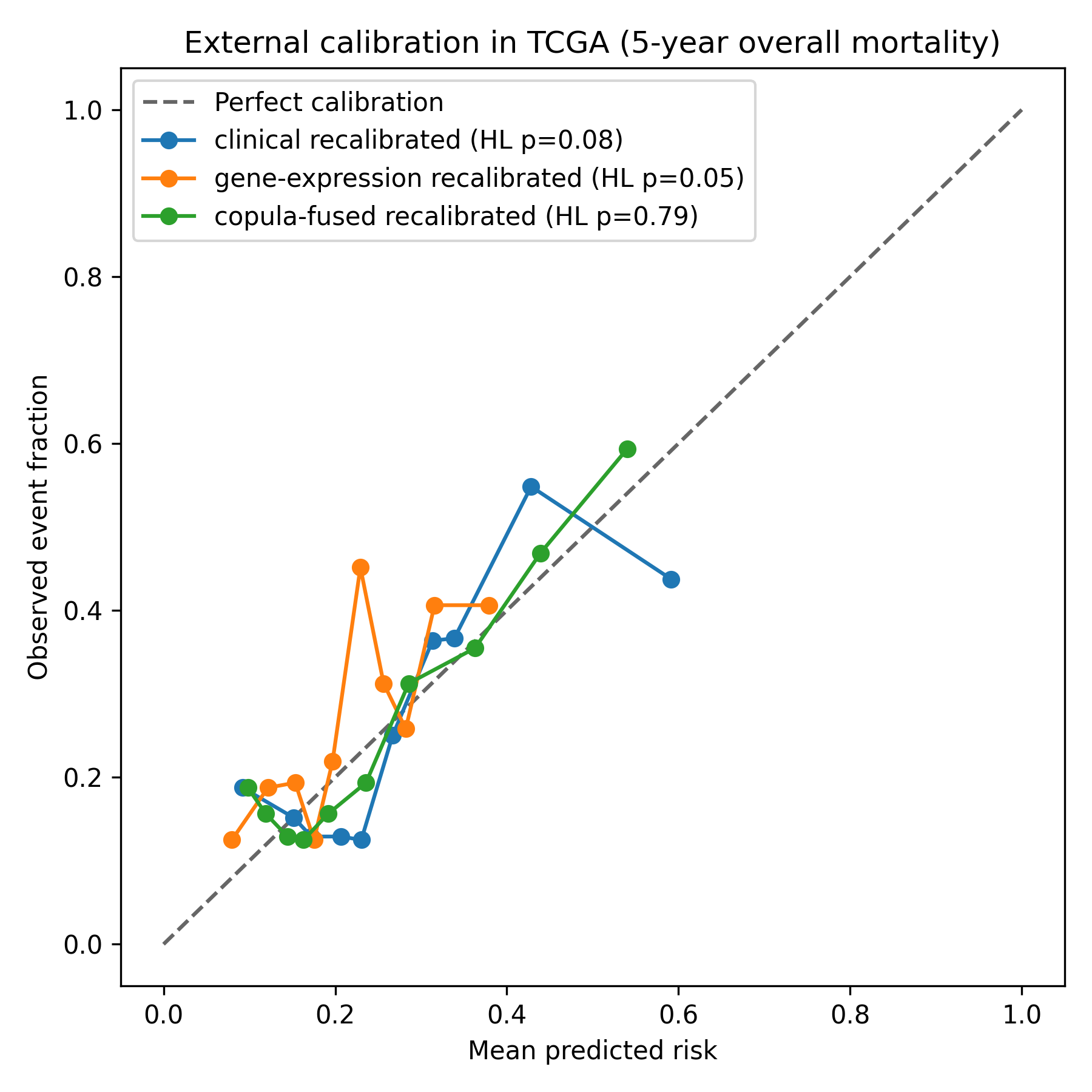}
\caption{Reliability curves for the identically recalibrated clinical,
gene-expression, and copula-fused scores in the internal METABRIC cohort (left)
and the external TCGA cohort (right). The dashed line denotes perfect
calibration.}
\label{fig:calibration}
\end{figure}

\subsection*{Feature Importance and  Exploratory Biological Context}

To improve interpretability, we examined which predictors most influenced each view using repeated five-fold cross-validated
permutation importance. Importance was defined as the mean reduction in held-out ROC-AUC across $25$ estimates per feature,
obtained from five folds and five permutations per fold. Tree impurity importance was retained only as a secondary descriptive
quantity. A feature was classified as \emph{stable-positive} when its empirical $2.5$th percentile was above zero and at least 80\% of its
estimates were positive.

% To improve interpretability, we examined which predictors most influenced each
% view, ranking features by repeated 5-fold cross-validated permutation importance
% (mean reduction in held-out ROC-AUC, $25$ estimates per feature from $5$ folds
% $\times$ $5$ repeats), which measures a feature's out-of-sample predictive
% contribution. Tree impurity importance was retained only as a secondary
% descriptive quantity. For each feature, we also recorded whether its importance
% was \emph{stable-positive}, defined as a $2.5$th empirical percentile above zero
% together with a positive importance in at least $80\%$ of the estimates.

In the clinical view, the highest-ranked predictors were the Nottingham
Prognostic Index, tumor size, age at diagnosis, number of positive lymph nodes,
and estrogen- and progesterone-receptor status, followed by the molecular
subtype labels (PAM50 and three-gene classifier) and tumor stage
(Table~\ref{tab:clinical_importance}). This ordering recapitulates classical
breast cancer prognostic factors and provides a sanity check on the clinical
model. Under the stringent stability criterion, however, only age at diagnosis
was stable-positive (positive in $96\%$ of estimates); the remaining predictors,
including the top-ranked Nottingham Prognostic Index, showed positive but more
variable importance across folds, reflecting the small number of clinical
predictors and their mutual correlation.

\begin{figure}[ht]
\centering
\includegraphics[width=0.6\textwidth]{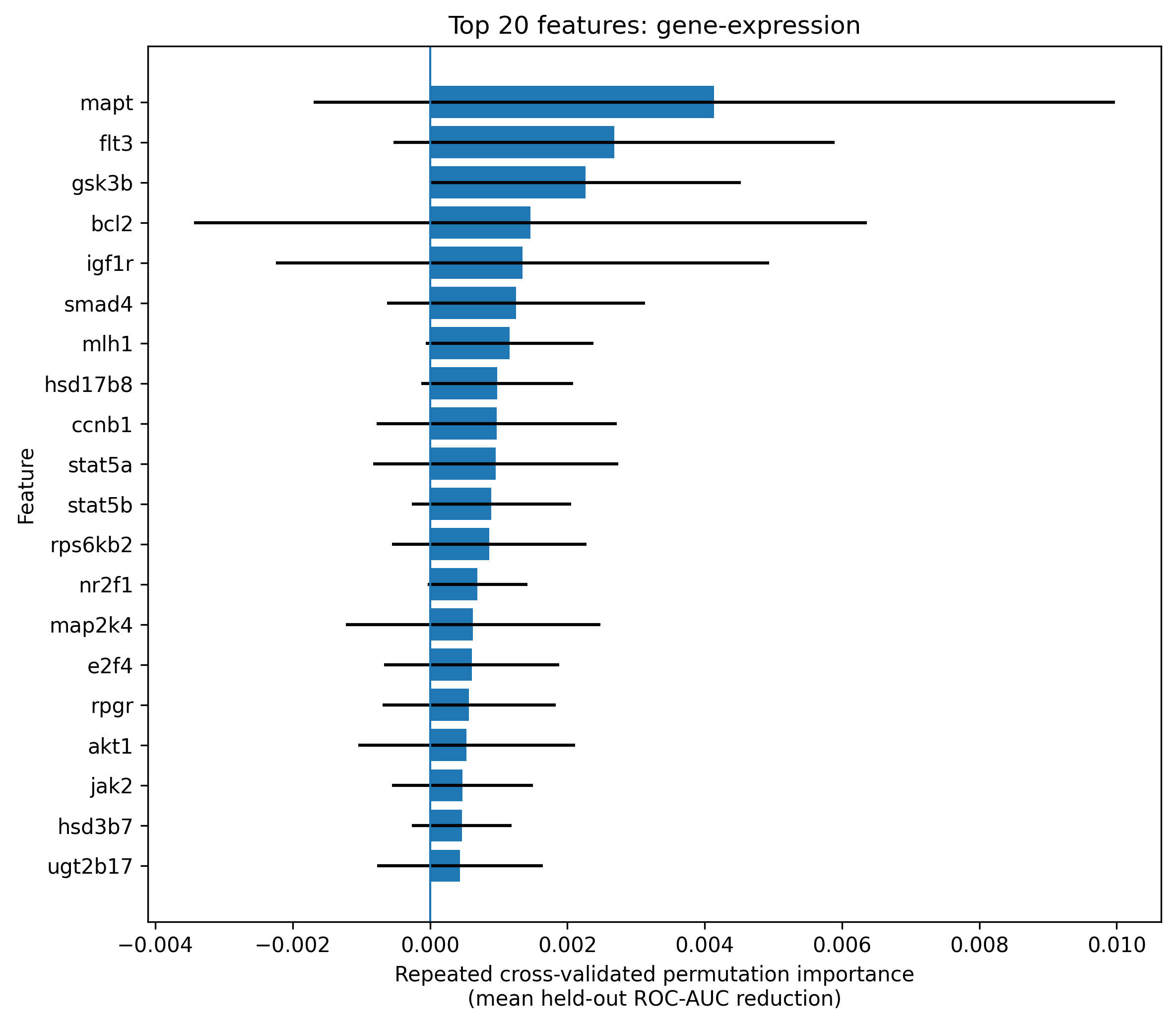}
\caption{Top twenty gene-expression features by repeated cross-validated
permutation importance in the primary METABRIC analysis. Bars show the mean
held-out ROC-AUC reduction; error bars show one standard deviation across the
$25$ permutation estimates.}
\label{fig:gene_importance}
\end{figure}

For the gene-expression analysis, Table~\ref{tab:gene_importance} lists the twenty highest-ranked genes according to repeated cross-validated permutation importance, and Figure~\ref{fig:gene_importance} presents the corresponding ranking. No individual gene met the prespecified stability criterion. The estimated importance values were small, and the empirical 2.5th percentile was below zero for every gene. The permutation-importance and random-forest impurity rankings also showed only weak agreement, with Spearman's rank correlation of $\rho = 0.14$ ($p = 0.002$) and a top-20 overlap of $8/20$. These findings indicate that the exact gene ordering was sensitive to the importance measure. Therefore, the gene-level results should be regarded as exploratory and should not be interpreted as identifying a stable set of predictive drivers.

The highest-ranked genes included \textit{MAPT}, \textit{FLT3}, \textit{GSK3B}, \textit{BCL2}, \textit{IGF1R}, \textit{SMAD4}, \textit{MLH1}, \textit{HSD17B8}, \textit{CCNB1}, \textit{STAT5A}, and \textit{STAT5B}. Published studies provide biological context for selected genes in this list. \textit{MAPT} is regulated by estrogen and has been investigated as a marker of paclitaxel sensitivity in breast cancer \cite{rouzier2005microtubule}. \textit{BCL2} expression has been associated with favorable prognosis across breast cancer molecular subtypes \cite{dawson2010bcl2}. \textit{IGF1R} is part of a growth-factor signaling pathway that has been implicated in therapeutic resistance \cite{casa2008type}. \textit{SMAD4} is a canonical mediator of transforming growth factor beta, TGF-$\beta$, signaling, a pathway with context-dependent tumor-suppressive and tumor-promoting roles in cancer \cite{massague2008tgfbeta}. \textit{CCNB1} and \textit{E2F4} participate in cell-cycle regulation \cite{malumbres2009cell,nevins2001rb}.

These citations establish biological plausibility only. They do not demonstrate that the listed genes are stable predictors, independent prognostic factors, causal drivers, or confirmed pathway-level mechanisms in the present analysis. No pathway-enrichment analysis was performed.

\begin{table}[ht]
\centering
\caption{Top clinical predictors by repeated cross-validated permutation
importance in the primary METABRIC analysis. ``Mean'' is the mean decrease in
ROC-AUC over $25$ estimates; the range is the empirical $2.5$th--$97.5$th
percentile of those estimates (which are not independent and so should not be
read as a confidence interval); ``Pos.\ frac.'' is the fraction of estimates
above zero. A predictor is \emph{stable-positive} when the $2.5$th percentile is
above zero and the positive fraction is at least $0.80$. Only age at diagnosis
met this criterion.}
\label{tab:clinical_importance}
\small
\begin{tabular}{lcccc}
\hline
Clinical predictor & Mean & $2.5$--$97.5\%$ range & Pos.\ frac. & Stable \\
\hline
Nottingham Prognostic Index   & 0.0197 & [$-0.0016$, $0.0373$] & 0.92 & No \\
Tumor size                   & 0.0155 & [$-0.0034$, $0.0343$] & 0.88 & No \\
Age at diagnosis              & 0.0133 & [$0.0017$, $0.0250$]  & 0.96 & Yes \\
Positive lymph nodes          & 0.0119 & [$-0.0023$, $0.0213$] & 0.92 & No \\
ER status                     & 0.0059 & [$-0.0079$, $0.0152$] & 0.76 & No \\
PR status                     & 0.0050 & [$-0.0099$, $0.0161$] & 0.84 & No \\
PAM50 + claudin-low subtype   & 0.0047 & [$-0.0082$, $0.0174$] & 0.64 & No \\
Three-gene classifier subtype & 0.0039 & [$-0.0103$, $0.0158$] & 0.68 & No \\
Tumor stage                  & 0.0034 & [$-0.0035$, $0.0108$] & 0.88 & No \\
Primary tumor laterality     & 0.0032 & [$-0.0036$, $0.0117$] & 0.80 & No \\
\hline
\end{tabular}
\end{table}

\begin{table}[ht]
\centering
\caption{Top twenty gene-expression features by repeated cross-validation
permutation importance in the primary METABRIC analysis. ``Mean'' is the mean
held-out ROC-AUC reduction over $25$ estimates; the range is the empirical
$2.5$th--$97.5$th percentile of those estimates (not a confidence interval, as
the estimates are not independent); ``Pos.\ frac.'' is the fraction of estimates
above zero. No gene met the stability criterion ($2.5$th percentile above zero
and positive fraction at least $0.80$).}
\label{tab:gene_importance}
\small
\begin{tabular}{clcccc}
\hline
Rank & Gene & Mean & $2.5$--$97.5\%$ range & Pos.\ frac. & Stable \\
\hline
1  & \textit{MAPT}    & 0.00414 & [$-0.0082$, $0.0138$] & 0.88 & No \\
2  & \textit{FLT3}    & 0.00268 & [$-0.0031$, $0.0077$] & 0.76 & No \\
3  & \textit{GSK3B}   & 0.00226 & [$-0.0021$, $0.0054$] & 0.84 & No \\
4  & \textit{BCL2}    & 0.00146 & [$-0.0058$, $0.0090$] & 0.56 & No \\
5  & \textit{IGF1R}   & 0.00135 & [$-0.0053$, $0.0070$] & 0.64 & No \\
6  & \textit{SMAD4}   & 0.00125 & [$-0.0018$, $0.0044$] & 0.72 & No \\
7  & \textit{MLH1}    & 0.00116 & [$-0.0012$, $0.0030$] & 0.84 & No \\
8  & \textit{HSD17B8} & 0.00098 & [$-0.0007$, $0.0033$] & 0.84 & No \\
9  & \textit{CCNB1}   & 0.00097 & [$-0.0016$, $0.0048$] & 0.68 & No \\
10 & \textit{STAT5A}  & 0.00096 & [$-0.0022$, $0.0038$] & 0.72 & No \\
11 & \textit{STAT5B}  & 0.00089 & [$-0.0017$, $0.0024$] & 0.84 & No \\
12 & \textit{RPS6KB2} & 0.00087 & [$-0.0017$, $0.0030$] & 0.80 & No \\
13 & \textit{NR2F1}   & 0.00069 & [$-0.0005$, $0.0022$] & 0.72 & No \\
14 & \textit{MAP2K4}  & 0.00063 & [$-0.0028$, $0.0032$] & 0.72 & No \\
15 & \textit{E2F4}    & 0.00061 & [$-0.0014$, $0.0027$] & 0.64 & No \\
16 & \textit{RPGR}    & 0.00057 & [$-0.0018$, $0.0026$] & 0.72 & No \\
17 & \textit{AKT1}    & 0.00053 & [$-0.0015$, $0.0037$] & 0.64 & No \\
18 & \textit{JAK2}    & 0.00047 & [$-0.0013$, $0.0022$] & 0.64 & No \\
19 & \textit{HSD3B7}  & 0.00046 & [$-0.0008$, $0.0017$] & 0.76 & No \\
20 & \textit{UGT2B17} & 0.00044 & [$-0.0013$, $0.0027$] & 0.60 & No \\
\hline
\end{tabular}
\end{table}

In the next section, we summarise the main findings of this study and outline directions for future work.
%%%%%%%%%%%%%%%%%%%%%%%%%%%%%%%%%%%%%%%%%%%%%%%%%%%
\section*{Conclusion and Future Work}
\label{sec:con}

In this study, we examined whether combining clinical and gene-expression machine learning risk scores can provide useful additional structure for breast cancer risk stratification beyond either view alone. Using the METABRIC breast cancer cohort, we built separate clinical and gene-expression models for predicting 5-year cancer-specific death. The clinical model showed the strongest discrimination in the primary METABRIC analysis (AUC approximately 0.78), while the gene-expression model provided a moderate but still informative predictive signal (AUC approximately 0.72). These findings suggest that the two views capture overlapping but not identical aspects of disease severity.

We then modeled the dependence between the clinical and gene-expression risk scores using several copula families. In the METABRIC cancer-specific analysis, the Frank copula provided the best fit among the candidate models considered, while the Gaussian copula performed very similarly. This indicates that the two scores were positively related, with a moderate overall dependence structure, but without strong evidence that a heavily tail-dependent model was necessary to describe the main joint pattern. Although the copula-fused score did not exceed the clinical score in ROC-AUC in the primary METABRIC analysis, the copula framework remained useful for describing the joint behavior of the two views and for defining descriptively interpretable median-defined joint score groups.

When we grouped patients by whether their clinical and gene-expression scores were above or below the median, the resulting strata showed clear differences in long-term outcomes. The low--low group had the most favorable survival profile, the discordant groups had intermediate outcomes, and the high-both group had the least favorable outcomes during follow-up. This pattern was also supported by the competing-risks analysis, in which the cumulative incidence of cancer death was highest in the high-both group. Taken together, these results suggest that joint clinical and gene-expression risk patterns may help identify subgroups with poorer prognosis, even when simple fusion does not improve discrimination over the strongest individual model. To examine generalizability, we also performed an external evaluation in an independent TCGA cohort using a harmonized pipeline based on shared clinical predictors and shared gene-expression features. In the harmonized TCGA evaluation, the copula-fused score showed discrimination comparable to the individual and simple-fusion scores. Its external calibration point estimates were also favourable after the same METABRIC-based recalibration procedure was applied to all three scores. In the paired Brier-score analysis, the copula-fused score had a significantly lower Brier score than the gene-expression score. By contrast, its Brier-score difference from the clinical score did not reach the conventional \(0.05\) significance level. The evidence for improved external predictive accuracy was therefore specific to the comparator. It supports a lower overall prediction error relative to the gene-expression score but does not establish a general advantage over both component scores. The magnitude of any benefit is likely to depend on the degree of alignment in available predictors, endpoint definitions, measurement platforms, and cohort structure. The identifiable contribution of the copula remains its explicit parametric characterization of score dependence and its structured descriptive representation of the joint score distribution.

This work has several limitations that point directly to future extensions. First, the primary METABRIC analysis and the external analysis did not use exactly the same endpoint or predictor set. The METABRIC cancer-specific analysis used a richer clinical view and a cancer-specific 5-year endpoint, whereas the harmonized TCGA analysis necessarily relied on a reduced shared clinical feature set (age at diagnosis and tumor stage), the 476 shared gene-expression features, and a 5-year overall-survival endpoint. This mismatch has a direct implication for how the external result should be interpreted: the harmonized analysis is best described as an external \emph{evaluation of the fusion approach} under a shared, deliberately reduced specification, rather than a direct external \emph{evaluation} of the specific richer clinical model developed in METABRIC. We have therefore avoided claiming that the primary model itself was evaluated out-of-cohort, and instead report that the copula-based fusion strategy retained favorable discrimination and, after recalibration, favorable calibration in an independent cohort. Because METABRIC used microarray data whereas TCGA used RNA-seq data, cohort-specific $z$-score normalization could not remove platform-specific measurement differences, batch effects, or shifts in expression distributions. We quantified the remaining distributional shift directly, obtaining a median absolute standardized mean difference of $0.066$, with $12$ of $476$ genes exceeding $1.0$. These differences may have contributed to the attenuation in transferred performance. Among the clinical, gene-expression, and fused scores, the fused score showed the most favorable external calibration point estimates in TCGA. The paired Brier-score comparison showed a lower Brier score than the gene-expression score, but the difference relative to the clinical score did not meet the conventional significance threshold. These comparisons condition on the fitted METABRIC models and recalibration functions and quantify uncertainty arising from the TCGA evaluation cohort. They do not propagate uncertainty from model selection, model fitting, copula estimation, or recalibration in METABRIC. The calibration findings should therefore be interpreted as comparator-specific and preliminary rather than as evidence of universal calibration superiority.
% Because METABRIC (microarray) and TCGA (RNA-seq) also differ in expression platform, with cohort-specific $z$-score normalization, some attenuation of transferred performance is expected; we quantified this shift directly (median absolute standardized mean difference $0.066$, with $12$ of $476$ genes above $1.0$). Against this background, the observation that, among the clinical, gene-expression, and fused scores, the fused score remained the best-calibrated in TCGA---and was competitive in discrimination, though not superior to simple score-level fusion---should be read as encouraging but preliminary. 
A second, related point concerns absolute-risk calibration: because the view-specific models were trained with balanced class weights to prioritize discrimination, their raw predicted probabilities were over-dispersed and, in the cancer-specific analysis, over-predicted absolute risk; an identical cross-fitted recalibration applied to all three scores substantially improved calibration, although the harmonized internal copula-fused score retained residual miscalibration by the HL test, and such a step would be required before any predicted probabilities were used for individual-level risk communication. Third, our machine learning models were restricted to regularized logistic regression and tree-based ensemble methods. More flexible survival-aware approaches, such as random survival forests, boosting methods for time-to-event outcomes, or neural survival models, could be explored in future work, provided that out-of-sample risk scores remain available for copula-based dependence modeling. Fourth, our copula analysis focused on combining one clinical score and one gene-expression score. In practice, clinicians may wish to integrate additional views, such as imaging, pathology, or treatment-specific risk estimates. Extending the framework to vine or factor copulas could support broader multi-view modeling while preserving the separation between marginal prediction and dependence structure.
In the present study we compared the copula-based fusion against simple score-level baselines (rank averaging, probability averaging, and cross-validated logistic stacking) and found that it did not improve discrimination over them; future work should extend this comparison to combined-feature models and to more expressive dependence-aware fusion schemes (for example vine or factor copulas, or copula constructions coupled to a supervised objective) to clarify the settings, if any, in which dependence-aware fusion provides a discrimination advantage rather than only interpretability and stratification benefits. Comparison with deep learning approaches and extension to additional data modalities, such as imaging or pathology, could also be explored in future work, particularly in higher-dimensional multimodal settings.

A further direction is to study whether the clinical--gene-expression dependence structure varies across clinically relevant subgroups, such as molecular subtype, treatment category, or stage. In addition, repeated evaluation across multiple external cohorts with more closely harmonized endpoints would help clarify when copula-based fusion offers a consistent benefit in discrimination and when its main value lies instead in patient stratification. This study focuses on the statistical and methodological aspects of copula-based score fusion; the resulting joint risk patterns should therefore be interpreted primarily in that context, while broader clinical interpretation may benefit from appropriate domain expertise.

In summary, this study shows that copula-based modeling provides a principled approach to examining the joint structure of clinical and gene-expression machine-learning risk scores in breast cancer. In METABRIC, copula modelling provided an explicit description of dependence and supported descriptive joint score-group analyses.
In the harmonized TCGA evaluation, the fused score showed measurable discrimination but did not clearly outperform the individual or
simple-fusion scores. These findings support further methodological evaluation of copula-based score fusion but do not establish
predictive superiority, direct validation of the primary model, or clinical utility.

% In the primary METABRIC analysis, the framework was especially useful for dependence modeling and joint risk stratification, while in the harmonized TCGA evaluation, it showed a modest but favorable external discrimination pattern for the fused score. Overall, these findings support copula-based fusion as a useful and interpretable approach for studying multimodal prognostic scores in breast cancer.
%%%%%%%%%%%%%%%%%%%%%%%%%%%%%%%%%%%%%%%%%%%%%%%%%%

\section*{Data availability}
The METABRIC cohort was originally generated and described by Curtis et al.~\cite{curtis2012} and Pereira et al.~\cite{pereira2016}; we used the curated, pre-processed version distributed publicly on Kaggle~\cite{metabricKaggle} (\url{https://www.kaggle.com/datasets/raghadalharbi/breast-cancer-gene-expression-profiles-metabric}). The TCGA external evaluation dataset was constructed from the cBioPortal study \textit{Breast Invasive Carcinoma (TCGA, PanCancer Atlas)}, available at \url{https://www.cbioportal.org/study/summary?id=brca_tcga_pan_can_atlas_2018}.

\section*{Code availability}
All code to reproduce the analyses is available at\\
\url{https://github.com/agnivibes/copula-risk-fusion-gene-expression-ml}.

The repository contains separate scripts for the primary METABRIC analysis,
TCGA dataset construction, and harmonized METABRIC-to-TCGA external evaluation.
The primary METABRIC script generates the internal machine-learning risk scores,
copula fits, survival summaries, and figures; the TCGA preprocessing script
constructs the harmonized evaluation dataset; and the external-evaluation script
trains the harmonized METABRIC models and evaluates the clinical, gene-expression,
and copula-fused scores in TCGA. All analyses used a fixed random seed ($42$);
gradient-boosted models (XGBoost) were run with GPU acceleration, and the full
software environment (package versions and the confirmed compute device) is
recorded in the \texttt{run\_environment.json} output of each script to support
reproducibility.

%%%%%%%%%%%%%%%%%%%%%%%%%%%%%%%%%%%%%%%%%%%%%%%%%%%%%%%%%
\bibliographystyle{plainnat}
\bibliography{refs}

%%%%%%%%%%%%%%%%%%%%%%%%%%%%%%%%%%%%%%%%%%%%%%%%%%%%%%%

\section*{Author contributions}
A.A. conceived and designed the study, developed the copula-based and machine learning methodology, implemented the analyses and code, ran the experiments, and interpreted the machine learning and methodological findings. S.H. supported the literature review, including genomics-related background research, assisted with the survival analysis, and contributed to the revision and final editing of the manuscript. M.M.M. contributed expertise in biostatistics and survival analysis, supported dataset selection, assisted with the construction of the TCGA external evaluation dataset, contributed to the introduction and related work sections write-up, compiled and organized the bibliography and citations, and helped interpret the results from a biostatistical data-analysis perspective.

\end{document}